\documentclass[10pt,twocolumn,letterpaper]{article}
\usepackage{cvpr}              

\usepackage{times}
\usepackage{epsfig}
\usepackage{graphicx}
\usepackage{amsmath}
\usepackage{amssymb}

\usepackage{multirow} 
\usepackage{rotating}

\usepackage{makecell} 
\usepackage[table,dvipsnames,xcdraw]{xcolor} 
\usepackage{colortbl} 
\usepackage{diagbox}  
\usepackage{siunitx}
\usepackage{pifont}
\usepackage[font=small]{caption}

\usepackage{enumitem} 
\usepackage{arydshln} 
\usepackage{tabularx}
\usepackage{booktabs} 
\usepackage{xspace}   
\usepackage[subrefformat=parens,labelformat=parens]{subcaption} 
\usepackage[Export]{adjustbox} 
\usepackage{pifont}   
\usepackage{balance}  

\usepackage{algorithm}
\usepackage{algpseudocode}

\usepackage{tikz}

\usetikzlibrary{arrows.meta, positioning}

\definecolor{cvprblue}{rgb}{0.21,0.49,0.74}
\usepackage[pagebackref,breaklinks,colorlinks,citecolor=cvprblue]{hyperref}

\usepackage[capitalize]{cleveref}

\def\ie{\emph{i.e.}\xspace}

\newcommand{\ours}{GS-SLAM\xspace}
\newcommand{\boldparagraph}[1]{\vspace{0.1em}\noindent{\bf #1}}

\newcolumntype{L}[1]{>{\raggedright\let\newline\\\arraybackslash\hspace{0pt}}m{#1}}
\newcolumntype{C}[1]{>{\centering\let\newline\\\arraybackslash\hspace{0pt}}m{#1}}
\newcolumntype{R}[1]{>{\raggedleft\let\newline\\\arraybackslash\hspace{0pt}}m{#1}}

\usepackage{multimedia}
\definecolor{lightyellow}{RGB}{255,255,224}

\usepackage[bigfiles]{pdfbase}
\ExplSyntaxOn
\NewDocumentCommand\embedvideo{smm}{
  \group_begin:
  \leavevmode
  \tl_if_exist:cTF{file_\file_mdfive_hash:n{#3}}{
    \tl_set_eq:Nc\video{file_\file_mdfive_hash:n{#3}}
  }{
    \IfFileExists{#3}{}{\GenericError{}{File~`#3'~not~found}{}{}}
    \pbs_pdfobj:nnn{}{fstream}{{}{#3}}
    \pbs_pdfobj:nnn{}{dict}{
      /Type/Filespec/F~(#3)/UF~(#3)
      /EF~<</F~\pbs_pdflastobj:>>
    }
    \tl_set:Nx\video{\pbs_pdflastobj:}
    \tl_gset_eq:cN{file_\file_mdfive_hash:n{#3}}\video
  }
  \pbs_pdfobj:nnn{}{dict}{
    /Type/RichMediaInstance/Subtype/Video
    /Asset~\video
    /Params~<</FlashVars (
    )>>
  }
  \pbs_pdfobj:nnn{}{dict}{
    /Type/RichMediaConfiguration/Subtype/Video
    /Instances~[\pbs_pdflastobj:]
  }
  \pbs_pdfobj:nnn{}{dict}{
    /Type/RichMediaContent
    /Assets~<<
      /Names~[(#3)~\video]
    >>
    /Configurations~[\pbs_pdflastobj:]
  }
  \tl_set:Nx\rmcontent{\pbs_pdflastobj:}
  \pbs_pdfobj:nnn{}{dict}{
    /Activation~<<
      /Condition/\IfBooleanTF{#1}{PV}{XA}
      /Presentation~<</Style/Embedded>>
    >>
    /Deactivation~<</Condition/PI>>
  }
  \hbox_set:Nn\l_tmpa_box{#2}
  \tl_set:Nx\l_box_wd_tl{\dim_use:N\box_wd:N\l_tmpa_box}
  \tl_set:Nx\l_box_ht_tl{\dim_use:N\box_ht:N\l_tmpa_box}
  \tl_set:Nx\l_box_dp_tl{\dim_use:N\box_dp:N\l_tmpa_box}
  \pbs_pdfxform:nnnnn{1}{1}{}{}{\l_tmpa_box}
  \pbs_pdfannot:nnnn{\l_box_wd_tl}{\l_box_ht_tl}{\l_box_dp_tl}{
    /Subtype/RichMedia
    /BS~<</W~0/S/S>>
    /Contents~(embedded~video~file:#3)
    /NM~(rma:#3)
    /AP~<</N~\pbs_pdflastxform:>>
    /RichMediaSettings~\pbs_pdflastobj:
    /RichMediaContent~\rmcontent
  }
  \phantom{#2}
  \group_end:
}
\ExplSyntaxOff

\colorlet{colorLow}{darkgray!30}            

\colorlet{colorFst}{Red!25}                 
\colorlet{colorSnd}{Orange!20}              
\colorlet{colorTrd}{Yellow!30}              
\colorlet{cmt}{darkgray!80}                 

\definecolor{R1}{HTML}{E97451}
\definecolor{R2}{HTML}{008080}
\definecolor{R3}{HTML}{0047AB}

\newcommand{\fs}{\cellcolor{colorFst}\bf}   
\newcommand{\nd}{\cellcolor{colorSnd}}      
\newcommand{\rd}{\cellcolor{colorTrd}}      

\definecolor{gray}{rgb}{0.65,0.65,0.65}
\definecolor{mycol}{rgb}{0.90,0.95,1.0}

\def\paperID{368}
\def\confName{CVPR}
\def\confYear{2024}

\title{GS-SLAM: Dense Visual SLAM with 3D Gaussian Splatting}
\author{
        Chi Yan$^{1,3}$\thanks{Equal contribution: \href{mailto:yanchi@pjlab.org.cn}{yanchi@pjlab.org.cn}.}  \hspace{3em} 
        Delin Qu$^{1,2}$\footnotemark[1] \hspace{3em} 
        Dan Xu$^{3}$ \hspace{3em}
        Bin Zhao$^{1,4}$ \\
        Zhigang Wang$^{1}$ \hspace{3em}
        Dong Wang$^{1}$\thanks{Corresponding author: \href{mailto:dongwang.dw93@gmail.com}{dongwang.dw93@gmail.com}.} \hspace{3em}
        Xuelong Li$^{1,5}$  \\
        $^{1}$Shanghai AI Laboratory \hspace{1em}
        $^{2}$Fudan University \hspace{1em}
        $^{3}$Hong Kong University of Science and Technology\\
        $^{4}$Northwestern Polytechnical University \hspace{4.5em}
        $^{5}$TeleAI, China Telecom Corp Ltd
}

\begin{document}
\maketitle

\begin{abstract}
In this paper, we introduce \textbf{GS-SLAM} that first utilizes  3D Gaussian representation in the Simultaneous Localization and Mapping (SLAM) system. It facilitates a better balance between efficiency and accuracy. Compared to recent SLAM methods employing neural implicit representations, our method utilizes a real-time differentiable splatting rendering pipeline that offers significant speedup to map optimization and RGB-D rendering. Specifically, we propose an adaptive expansion strategy that adds new or deletes noisy 3D Gaussians in order to efficiently reconstruct new observed scene geometry and improve the mapping of previously observed areas. This strategy is essential to extend 3D Gaussian representation to reconstruct the whole scene rather than synthesize a static object in existing methods. Moreover, in the pose tracking process, an effective coarse-to-fine technique is designed to select reliable 3D Gaussian representations to optimize camera pose, resulting in runtime reduction and robust estimation. Our method achieves competitive performance compared with existing state-of-the-art real-time methods on the Replica, TUM-RGBD datasets. Project page: \href{https://gs-slam.github.io/}{https://gs-slam.github.io/}.
\end{abstract}
\vspace{-3.0ex}
\section{Introduction}
\label{sec:intro}

\begin{figure}[t]
    \begin{center}
        \includegraphics[width=1\linewidth]{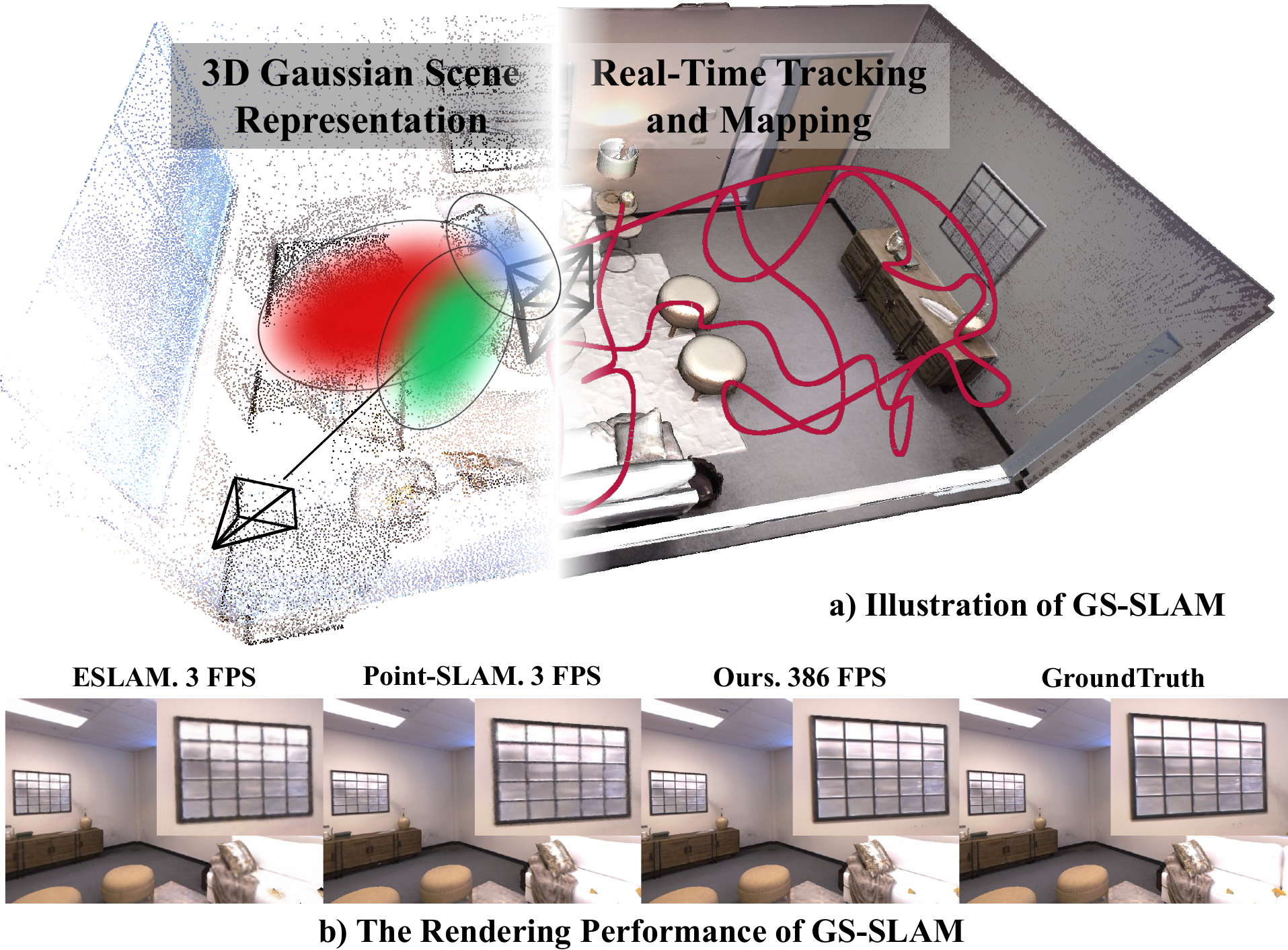}
    \end{center}
    \caption{The illustration of the proposed \ours. It first utilizes the 3D Gaussian representation and differentiable splatting rasterization pipeline in SLAM, achieving real-time tracking and mapping performance on GPU. Besides, benefiting from the splatting rasterization pipeline, \ours achieves a 100$\times$ faster rendering FPS and more high-quality full image results than the other SOTA methods.
    }
    \label{fig:head}
\end{figure}

Simultaneous localization and mapping (SLAM) has emerged as a pivotal technology in fields such as robotics~\cite{DurrantWhyte2006SimultaneousLA}, virtual reality~\cite{Jiang2022AS6}, and augmented reality~\cite{Reitmayr2010SimultaneousLA,Theodorou2022VisualSA}. The goal of SLAM is to construct a dense/sparse map of an unknown environment while simultaneously tracking the camera pose. Traditional SLAM methods employ point/surfel clouds~\cite{MurArtal2016ORBSLAM2AO,Wang2019RealtimeSD,whelan2015elasticfusion,Stckler2014MultiresolutionSM}, mesh representations~\cite{Ruetz2018OVPCM3}, voxel hashing~\cite{Niener2013Realtime3R,Maier2017EfficientOS,Khler2016HierarchicalVB} or voxel grids~\cite{Newcombe2011KinectFusionRD} as scene representations to construct dense mapping, and have made considerable progress on localization accuracy. However, these methods face serious challenges in obtaining fine-grained dense maps.

Recently, Neural Radiance Fields (NeRF)~\cite{mildenhall2020nerf} have been explored to enhance SLAM methodologies and exhibit strengths in generating high-quality, dense maps with low memory consumption~~\cite{Sucar2021iMAPIM}.
In particular, iMAP~\cite{Sucar2021iMAPIM} uses a single multi-layer perceptron (MLP) to represent the entire scene, which is updated globally with the loss between volume-rendered RGB-D image and ground-truth observations.
NICE-SLAM~\cite{Zhu2021NICESLAMNI} utilizes a hierarchical neural implicit grid as scene map representation to allow local updates for reconstructing large scenes.
Moreover, ESLAM~\cite{Johari2022ESLAMED}, CoSLAM~\cite{Wang2023CoSLAMJC} and EN-SLAM~\cite{qu2023implicit} utilize axis-aligned feature planes and joint coordinate-parametric encoding to improve the capability of scene representation, achieving efficient and high-quality surface map reconstruction. In practical mapping and tracking steps, these methods only render a small set of pixels to reduce optimization time, which leads to the reconstructed dense maps lacking the richness and intricacy of details. In essence, it is a trade-off for the efficiency and accuracy of NeRF-based SLAM since obtaining high-resolution images with the ray-based volume rendering technique is time-consuming and unacceptable. 

Fortunately, recent work~\cite{kerbl3Dgaussians,Luiten2023Dynamic3G,Wu20234DGS} with 3D Gaussian representation and tile-based splatting techniques has shown great superiority in the efficiency of high-resolution image rendering. It is applied to synthesize novel view RGB images of static objects, achieving state-of-the-art visual quality for 1080p resolution at real-time speed. Inspired by this, we extend the rendering superiority of 3D Gaussian scene representation and real-time differentiable splatting rendering pipeline for the task of dense RGB-D SLAM and manage to jointly promote the speed and accuracy of NeRF-based dense SLAM, as shown in~\cref{fig:head}.


To this end, we propose GS-SLAM, the first RGB-D dense SLAM system that first utilizes 3D Gaussian scene representation coupled with the splatting rendering technique to better balance speed and accuracy. Our system optimizes camera tracking and mapping with a novel RGB-D rendering approach that processes 3D Gaussians quickly and accurately through sorting and $\alpha$-blending. We enhance scene reconstruction by introducing an adaptive strategy for managing 3D Gaussian elements, which optimizes mapping by focusing on current observations and minimizes errors in dense maps and images. Moreover, we propose a coarse-to-fine approach, starting with low-resolution image analysis for initial pose estimation and refining it with high-resolution rendering using select 3D Gaussians, to boost speed and accuracy. We perform extensive evaluations on a selection of indoor RGB-D datasets and demonstrate state-of-the-art performance on dense neural RGB-D SLAM in terms of tracking, rendering, and mapping. Overall, our contributions include:

\begin{itemize}[itemsep=0pt,topsep=0pt,leftmargin=10pt]
    \item We propose GS-SLAM, the first 3D Gaussian Splatting(3DGS)-based dense RGB-D SLAM approach, which takes advantage of the fast splatting rendering technique to boost the mapping optimizing and pose tracking, achieving real-time and photo-realistic reconstruction performance.
    \item We present an adaptive 3D Gaussian expansion strategy to efficiently reconstruct new observed scene geometry and develop a coarse-to-fine technique to select reliable 3D Gaussians to improve camera pose estimation.
    \item Our approach achieves competitive performance on Replica and TUM-RGBD datasets in terms of tracking, and mapping and runs at 8.43 FPS, resulting in a better balance between efficiency and accuracy.
\end{itemize}


\section{Related Work}

\label{sec:relate}
\begin{figure*}[t]
    \begin{center}
        \includegraphics[width=1\linewidth]{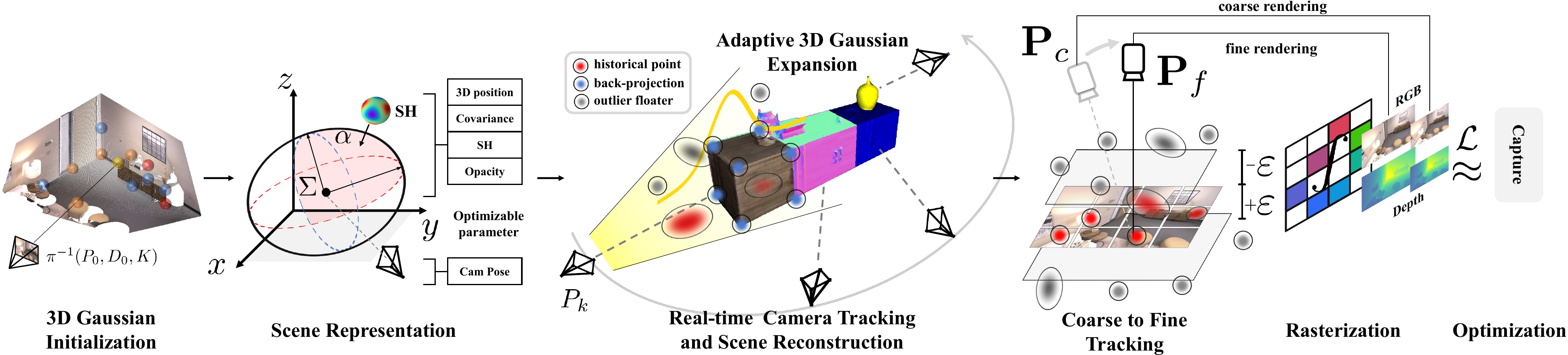}
    \end{center}
    \caption{Overview of the proposed method. We aim to use 3D Gaussians to represent the scene and use the rendered RGB-D image for inverse camera tracking. GS-SLAM proposes a novel Gaussian expansion strategy to make the 3D Gaussian feasible to reconstruct the whole scene and can achieve real-time tracking, mapping, and rendering performance on GPU.}
    \label{fig:pipeline}
\end{figure*}

\boldparagraph{Dense Visual SLAM.}
The existing real-time dense visual SLAM systems are typically based on discrete handcrafted features or deep-learning embeddings, and follow the mapping and tracking architecture in~\cite{Klein2009ParallelTA}. DTAM~\cite{Newcombe2011DTAMDT} first introduces a dense SLAM system that uses photometric consistency to track a handheld camera and represent the scene as a cost volume. KinectFusion~\cite{Whelan2012KintinuousSE} performs camera tracking by iterative-closest-point and updates the scene via TSDF-Fusion. BAD-SLAM~\cite{Schps2019BADSB} proposes to jointly optimize the keyframe poses and 3D scene geometry via a direct bundle adjustment (BA) technique. In contrast, recent works integrate deep learning with the traditional geometry framework for more accurate and robust camera tracking and mapping, such as DROID-SLAM~\cite{Teed2021DROIDSLAMDV}, CodeSLAM~\cite{Bloesch2018CodeSLAML}, SceneCode~\cite{Zhi2019SceneCodeMD}, and NodeSLAM~\cite{Sucar2020NodeSLAMNO}, have made significant advances in the field, achieving more accurate and robust camera tracking and mapping performance.

\boldparagraph{Neural Implict Radiance Field based SLAM.}
For NeRF-based SLAM, existing methods can be divided into three main types: \textit{MLP-based methods}, \textit{Hybrid representation methods}, and \textit{Explicit methods}.
MLP-based method iMAP~\cite{Sucar2021iMAPIM} offers scalable and memory-efficient map representations but faces challenges with catastrophic forgetting in larger scenes. Hybrid representation methods combine the advantages of implicit MLPs and structure features, significantly enhancing the scene scalability and precision. For example, NICE-SLAM~\cite{Zhu2021NICESLAMNI} integrates MLPs with multi-resolution voxel grids, enabling large scene reconstruction, and Vox-Fusion~\cite{Yang2022VoxFusionDT} employs octree expansion for dynamic map scalability, while ESLAM~\cite{Johari2022ESLAMED} and Point-SLAM~\cite{pointslam} utilize tri-planes and neural point clouds respectively to improve the mapping capability. As for the explicit method proposed in~\cite{Teigen2023RGBDMA}, it stores map features in voxel directly, without any MLPs, enabling faster optimization. Instead of representing maps with implicit features, \ours utilizes the 3D Gaussian representation, efficiently renders images using splatting-based rasterization, and optimizes parameters directly with backward propagation.

\boldparagraph{3D Gaussian Representation.}  Several recent approaches have use 3D Gaussians for  shape reconstruction, such as Fuzzy Metaballs~\cite{keselman2022fuzzy,keselman2023flexible}, VoGE~\cite{wang2022voge}, 3DGS~\cite{kerbl3Dgaussians}. Notably,  3DGS~\cite{kerbl3Dgaussians} demonstrates great superiorities in high-quality real-time novel-view synthesis. This work represents the scene with 3D Gaussians and develops a NeRF-style fast rendering algorithm to support anisotropic splatting, achieving SOTA visual quality and fast high-resolution rendering performance.  Beyond the rendering superiorities, Gaussian splatting holds an explicit geometry scene structure and appearance, benefiting from the exact modeling of scenes representation~\cite{Yang2023RealtimePD}. This promising technology has been rapidly applied in several fields, including 3D generation~\cite{Chen2023Textto3DUG,Tang2023DreamGaussianGG,Yi2023GaussianDreamerFG}, dynamic scene modeling~\cite{Luiten2023Dynamic3G}\cite{Wu20234DGS}\cite{Yang2023Deformable3G}, and photorealistic drivable avatars~\cite{Zielonka2023Drivable3G}. However, currently, there is no research addressing camera pose estimation or real-time mapping using 3D Gaussian models due to the inherent limitations of the prime pipeline~\cite{kerbl3Dgaussians}, \ie, prerequisites of initialized point clouds or camera pose inputs~\cite{colmap2016}. In contrast, we derive the analytical derivative equations for pose estimation in the Gaussian representation and implement efficient CUDA optimization.
\section{Methodology}
\label{sec:method}
\cref{fig:pipeline} shows the overview of the proposed \ours. We aim to estimate the camera poses $\left\{\mathbf{P}_{i}\right\}_{i=1}^{N}$ of every frame and simultaneously reconstruct a dense scene map by giving an input sequential RGB-D stream $\left\{\mathbf{I}_{i}, \mathbf{D}_{i}\right\}_{i=1}^{M}$ with known camera intrinsic $\mathbf{K} \in \mathbb{R}^{3 \times 3}$. In~\cref{sec:3D splatting},  we first introduce 3D Gaussian as the scene representation $\mathbf{S}$ and the RGB-D render by differentiable splatting rasterization. With the estimated camera pose of the keyframe, in ~\cref{sec:mapping}, an adaptive expansion strategy is proposed to add new or delete noisy 3D Gaussians to efficiently reconstruct new observed scene geometry while improving the mapping of the previously observed areas. For camera tracking of every input frame, we derive an analytical formula for backward optimization with rendering RGB-D loss. We further introduce an effective coarse-to-fine technique to minimize rendering losses to achieve efficient and accurate pose estimation in ~\cref{sec:Tracking}.

\subsection{3D Gaussian Scene Representation}
\label{sec:3D splatting}
Our goal is to optimize a scene representation that captures the geometry and appearance of the scene, resulting in a detailed dense map and high-quality novel view synthesis. To do this, we model the scene as a set of 3D Gaussians coupled with opacity and spherical harmonics
\begin{equation}
    \mathbf{G} = \{ G_i:(\mathbf{X}_i, \mathbf{\Sigma}_i, \Lambda_i, \boldsymbol{Y}_i) | i=1,...,N\}.
    \label{eq:scene-rep}
\end{equation}
Each 3D Gaussian scene representation $G_i$ is defined by position $\mathbf{X}_i \in \mathbb{R}^3$, 3D covariance matrix $\mathbf{\Sigma}_i \in \mathbb{R}^{3 \times 3}$, opacity $\Lambda_i \in \mathbb{R}$ and 1-degree spherical harmonics ($\boldsymbol{Y}$) per color channel, a total of 12 coefficients for $\boldsymbol{Y}_i \in \mathbb{R}^{12}$. To reduce the learning difficulty of the 3D Gaussians~\cite{Zwicker2001EWAVS}, we parameterize the 3D Gaussian's covariance as:
\begin{equation}
    \mathbf{\Sigma} = \mathbf{RSS}^{T}\mathbf{R}^{T},
    \label{eq:covariance}
\end{equation}
where $\mathbf{S} \in \mathbb{R}^{3}$ is a 3D scale vector, $\mathbf{R} \in \mathbb{R}^{3\times3}$ is the rotation matrix, storing as a 4D quaternion.

\boldparagraph{Color and depth splatting rendering.}
With the optimized 3D Gaussian scene representation parameters, given the camera pose $\mathbf{P} = \{\mathbf{R}, \mathbf{t}\}$, the 3D Gaussians $G$ are projected into 2D image plane for rendering with:
\begin{equation}
    \mathbf{\Sigma^{\prime}} = \mathbf{JP}^{-1}\mathbf{\Sigma P}^{-T}\mathbf{J}^{T},
    \label{eq:covariance2d}
\end{equation}
where $\mathbf{J}$ is the Jacobian of the affine approximation of the projective function. After projecting 3D Gaussians to the image plane, the color of one pixel is rendered by sorting the Gaussians in depth order and performing front-to-back $\alpha$-blending rendering as follows:
\begin{equation}
    \hat{\mathbf{C}}=\sum_{i \in N} \mathbf{c}_i \alpha_i \prod_{j=1}^{i-1}\left(1-\alpha_j\right),
    \label{eq:color_render}
\end{equation}
where $\mathbf{c}_i$ represents the color of the $i$-th 3D Gaussian obtained by learned spherical harmonics coefficients $\boldsymbol{Y}$, $\alpha_i$ is the density computed by learned opacity $\Lambda_i$ and 2D Gaussian with covariance $\Sigma^{\prime}$. Similarly, the depth is rendered by

\begin{equation}
    \hat{D} = \sum_{i \in N} d_i \alpha_i \prod_{j=1}^{i-1}\left(1-\alpha_j\right),
    \label{eq:depth}
\end{equation}
where $d_{i}$ denotes the depth of the center of the $i$-th 3D Gaussian, which is obtained by projecting to $z$-axis in the camera coordinate.

\subsection{Adaptive 3D Gaussian Expanding Mapping}
\label{sec:mapping}
The 3D Gaussian scene representations are updated and optimized on each selected keyframe for stable mapping. Given the estimated pose of each selected keyframe, we first apply the proposed adaptive expansion strategy to add new or delete noisy 3D Gaussians from the whole scene representations to render RGB-D images with resolution $H \times W$, and then the updated 3D Gaussian scene representations are optimized by minimizing the geometric depth loss $\mathcal{L}_d$ and the photometric color loss $\mathcal{L}_\mathbf{c}$ to the sensor observation depth $D$ and color $\mathbf{C}$,
\begin{equation}
    \mathcal{L}_{\mathbf{c}}=\sum_{m=1}^{HW}\left|\mathbf{C}_m-\hat{\mathbf{C}}_m\right|,\enspace \mathcal{L}_{d}=\sum_{m=1}^{HW}\left|D_m-\hat{D}_m\right|.
    \label{eq:map_loss}
\end{equation}
The loss optimizes the parameters of all 3D Gaussians that contribute to the rendering of these keyframe images.

\boldparagraph{Adaptive 3D Gaussian Expansion Strategy.}
At the first frame of the RGB-D sequence, we first uniformly sample half pixels from a whole image with $H \times W$ resolution and back-projecting them into 3D points $\mathbf{X}$ with corresponding depth observation $D$. The 3D Gaussian scene representations are created by setting position as $\mathbf{X}$ and initializing zero degree Spherical Harmonics coefficients with RGB color $\mathbf{C}_i$. The opacities are set to pre-defined values, and the covariance is set depending on the spatial point density, \ie,
\begin{equation}
    \{ G_i=(\mathbf{P}_i, \mathbf{\Sigma}_{init}, \Lambda_{init}, \mathbf{C}_i)| i=1,...,M\},
    \label{eq:init-rep}
\end{equation}
where $M$ equals to ${HW}/{2}$. The 3D Gaussians are initialized and then optimized using the first RGB-D image with rendering loss. Note that only half of the pixels are used to initialize the scene, leaving space to conduct adaptive density control of Gaussians that splits large points into smaller ones and clones them with different directions to capture missing geometric details. 

\textbf{\emph{Adding step:}} To obtain a complete map of the environment, the 3D Gaussian scene representations should be able to model the geometry and appearance of newly observed areas. Specifically, at every keyframe, we add first rendered RGB-D images using historical 3D Gaussians and calculate cumulative opacity $T=\sum_{i \in N} \alpha_i \prod_{j=1}^{i-1}\left(1-\alpha_j\right)$ for each pixel. We label one pixel as un-reliable $x^{un}$ if its cumulative opacity $T$ is too low or its rendered depth $\hat{D}$ is far away from observed depth $D$, \ie,
\begin{equation}
    \vspace{-1ex}
    T < \tau_T\quad  \text{or}\quad |D - \hat{D}| > \tau_D.
    \label{eq:add_method}
\end{equation}
These selected un-reliable pixels mostly capture new observed areas. Then we back-project these un-reliable pixels to 3D points $\mathbf{P}^{un}$, and a set of new 3D Gaussians at $\mathbf{P}^{un}$ initialized as Eq.~\ref{eq:init-rep} are added into scene representations to model the new observed areas.

\begin{figure}[t]
    \vspace{-1ex}
    \begin{center}
        \includegraphics[width=0.9\linewidth]{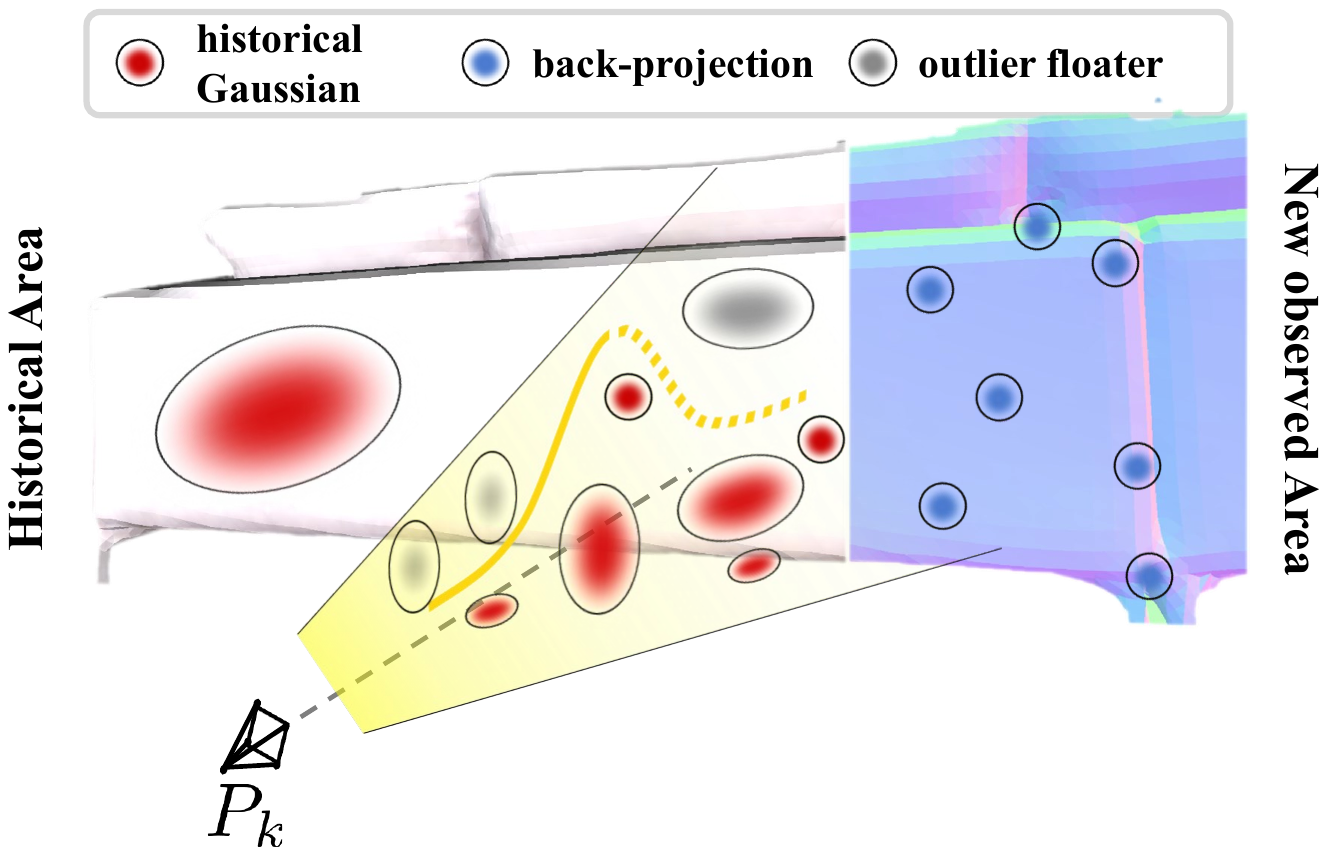}
    \end{center}
    \vspace{-2ex}
    \caption{Illustration of the proposed adaptive 3D Gaussian expansion strategy. \ours inhibits the low-quality 3D Gaussian floaters in the current frustum according to depth.}
    \vspace{-1ex}
    \label{fig:add_delete}
\end{figure}

\textbf{\emph{Deleting step:}} As shown in \cref{fig:add_delete}, there are some floating 3D Gaussians due to the unstable adaptive control of Gaussians after optimization with Eq.~\ref{eq:map_loss}. These floating 3D Gaussians will result in a low-quality dense map and a rendered image containing lots of artifacts. To address this issue, after adding new 3D Gaussians, we check all visible 3D Gaussians in the current camera frustum and significantly decrease opacity $\Lambda_i$ of 3D Gaussians whose position is not near the scene surfaces. Formally, for each visible 3D Gaussian, we draw a ray $\mathbf{r}(t)$ from camera origin $\mathbf{o}$ and its position $\mathbf{X}_i=(x_i,y_i,z_i)$, \ie, $\mathbf{r}(t)=\mathbf{o}+t(\mathbf{X}_i-\mathbf{o})$. Then, we can find a pixel with coordinate $(u,v)$ where this ray intersects the image plane and corresponding depth observation $D$. The 3D Gaussians are deleted by degenerating its opacity as follows:
\begin{equation}
    \vspace{-1ex}
    G_i:\Lambda_i \Rightarrow G_i:\eta \Lambda_i, \ \text{if} \ D-dist(\mathbf{X}_i, \mathbf{P}_{uv}) > \gamma,
    \label{eq:delete_method}
\end{equation}
where $P_{uv}$ is the world coordinates of the intersected pixel calculated with the camera intrinsic and extrinsic. $dist(\cdot,\cdot)$ is the Euclidean distance, and $\eta$ (much smaller than 1) and $\gamma$ are the hyper-parameters. Note that we decrease the opacity of floating 3D Gaussians in front of the scene surfaces to make our newly added 3D Gaussians well-optimized.

\subsection{Tracking and Bundle Adjustment}
\label{sec:Tracking}
In the parallel camera tracking phase of our work, we first employ a common straightforward constant velocity assumption to initialize new poses. This assumption transforms the last known pose based on the relative transformation between the second-to-last pose and the last pose. Then, the accurate camera pose $\mathbf{P}$ is optimized by minimizing rendered color loss, \ie,
\begin{equation}
    \mathcal{L}_{track}=\sum_{m=1}^{M}\left|\mathbf{C}_m-\hat{\mathbf{C}}_m\right|_1,
    \enspace \min_{\mathbf{R}, \mathbf{t}}(\mathcal{L}_{track}),
    \label{eq:track_loss}
\end{equation}
where $M$ is the number of sampled pixels for rendering.

\boldparagraph{Differentiable pose estimation.}
According to~\cref{eq:covariance2d,eq:color_render}, we observe that the gradient of the camera pose $\mathbf{P}$ is related to three intermediate variables: $\mathbf{\Sigma^{\prime}}$, $\mathbf{c}_{i}$, and the projected coordinate $\mathbf{m}_i$ of Gaussian $G_i$. By applying the chain rule of derivation, we obtain the analytical formulation of camera pose $\mathbf{P}$:
\begin{equation}\label{eq:pose_grad}
    \resizebox{0.87\linewidth}{!}{
        \begin{math}
            \begin{aligned}
                \frac{\partial {\mathcal{L_\mathbf{c}}}}{\partial{\mathbf{P}}} & = \frac{\partial{\mathcal{L_\mathbf{c}}}}{\partial{\mathbf{C}}} \frac{\partial{\mathbf{C}}}{\partial{\mathbf{P}}} = \frac{\partial{\mathcal{L_\mathbf{c}}}}{\partial{\mathbf{C}}} \left ({\frac{\partial{\mathbf{C}}}{\partial{\mathbf{c}_i}} \frac{\partial{\mathbf{c}_i}}{\partial{\mathbf{P}}}} + \frac{\partial{\mathbf{C}}}{\partial{\mathbf{\alpha}_i}} \frac{\partial{\mathbf{\alpha}_i}}{\partial{\mathbf{P}}}\right ) \\
                                                                & = \frac{\partial{\mathcal{L_\mathbf{c}}}}{\partial{\mathbf{C}}} \frac{\partial{\mathbf{C}}}{\partial{\mathbf{\alpha}_i}} \left ( \frac{\partial{\alpha_i}}{\partial{\mathbf{\Sigma^\prime}}} \frac{\partial\mathbf{\Sigma^\prime}}{\partial{\mathbf{P}}} + \frac{\partial{\alpha_i}}{\partial{\mathbf{m_i}}} \frac{\partial\mathbf{m_i}}{\partial{\mathbf{P}}} \right )                                            \\
                                                                & = \frac{\partial{\mathcal{L_\mathbf{c}}}}{\partial{\mathbf{C}}} \frac{\partial{\mathbf{C}}}{\partial{\mathbf{\alpha}_i}} \left ( \frac{\partial{\alpha_i}}{\partial{\mathbf{\Sigma^\prime}}} \frac{\partial({\mathbf{JP}^{-1}\mathbf{\Sigma P}^{-T}\mathbf{J}^{T})}}{\partial{\mathbf{P}}} + \frac{\partial \alpha_i}{\partial \mathbf{m_i}}\frac{\partial(\mathbf{KP}\mathbf{X}_i)}{\partial{\mathbf{P} d_i}}\right )
            \end{aligned}
        \end{math}
    },
\end{equation}
where $d_i$ denotes the $z$-axis coordinate of projection $\mathbf{m}_i$. The item ${\frac{\partial{\mathbf{C}}}{\partial {\mathbf{c}_i}} \frac{\partial{\mathbf{c}_i}}{\partial{\mathbf{P}}}}$ can be eliminated because we are only concerned about the view-independent color in our tracking implementation. In addition, we find that the intermediate gradient $\frac{\partial(\mathbf{KP}\mathbf{X}_i)}{\partial{\mathbf{P} d_i}}$ is the deterministic component for the camera pose $\mathbf{P}$. So we simply ignore the back-propagation of $\frac{\partial({\mathbf{JP}^{-1}\mathbf{\Sigma P}^{-T}\mathbf{J}^{T})}}{\partial{\mathbf{P}}}$ for efficiency. More details can be found in the supplemental materials.

\boldparagraph{Coarse-to-fine camera tracking.}
It would be problematic to optimize the camera pose with all image pixels since artifacts in images can cause drifted camera tracking. To address this issue, as shown in Fig.~\ref{fig:pipeline}, in the differentiable pose estimation step for each frame, we first take advantage of image regularity to render only a sparse set of pixels and optimize tracking loss to obtain a coarse camera pose. This coarse optimization step significantly eases the influence of detailed artifacts. Further, we use this coarse camera pose and depth observation to select reliable 3D Gaussians, which guides \ours to render informative areas with clear geometric structures to refine coarse camera pose via further optimizing tracking loss on new rendering pixels.

Specifically, in the coarse stage, we first render a coarse image $\hat{\mathbf{I}}_c$ with resolution $H/2 \times W/2$ at uniformly sampled image coordinates and optimize tracking loss in Eq.~\ref{eq:track_loss} for $T_c$ iterations, and the obtained camera pose is denoted as $\mathbf{P}_c$. In the fine stage, we use a similar technique with adaptive 3D Gaussian expansion strategy in Section~\ref{sec:mapping} to select reliable 3D Gaussian to render full-resolution images while ignoring noisy 3D Gaussians that cause artifacts. In detail, we check all visible 3D Gaussians under coarse camera pose $\mathbf{P}_c$, and remove 3D Gaussians whose position is far away from the scene surface. Formally, for each visible 3D Gaussians $G_i$ with position $\mathbf{X}_i$, we project it to the camera plane using coarse camera pose $\mathbf{P}_c$ and camera intrinsic. Given the projected pixel's depth observation $D_i$ and the distance $d_i$ that is between 3D Gaussians $G_i$ and the camera image plane, the reliable 3D Gaussians are selected as follows:
\begin{equation}
    \begin{split}
        \mathbf{G}_{selected} &= \{ G_i | G_i \in \mathbf{G}~\text{and}~abs(D_i-d_i) \leq \varepsilon\},\\
        \hat{\mathbf{I}}_f &= \mathcal{F}(u,v,\mathbf{G}_{selected}),
    \end{split}
    \label{eq:coarse2fine}
\end{equation}
where we use the selected reliable 3D Gaussians to render full-resolution images $\hat{\mathbf{I}}_f$. $u,v$ denote the pixel coordinates in $\hat{\mathbf{I}}_f$, and $\mathcal{F}$ represents the color splatting rendering function. The final camera pose $\mathbf{P}$ is obtained by optimizing tracking loss in Eq.~\ref{eq:track_loss} with $\hat{\mathbf{I}}_f$ for other $T_f$ iterations. Note that $\hat{\mathbf{I}}_c$ and $\hat{\mathbf{I}}_f$ are only rendered at previously observed areas, avoiding rendering areas where 3D scene representations have not been optimized in the mapping process. Also, we add keyframes based on the proportion of the currently observed image's reliable region to the overall image. At the same time, when the current tracking frame and most recent keyframe differ by more than a threshold value $\mu_{k}$, this frame will be inserted as a keyframe.

\boldparagraph{Bundle adjustment.}
In the bundle adjustment (BA) phase, we optimize the camera poses $\mathbf{P}$ and the 3D Gaussian scene representation $\mathbf{S}$ jointly. We randomly select $K$ keyframes from the keyframe database for optimization, using the loss function similar to the mapping part.
For pose optimization stability, we only optimize the scene representation $\mathbf{S}$ in the first half of the iterations. In the other half of the iterations, we simultaneously optimize the map and the poses. Then, the accurate camera pose $\mathbf{P}$ is optimized by minimizing rendering color loss, \ie,
\begin{equation}
    \resizebox{0.87\linewidth}{!}{
        \begin{math}
            \begin{aligned}
                \mathcal{L}_{ba}=\frac{1}{K}\sum_{k=1}^{K}\sum_{m=1}^{HW}\left|D_m-\hat{D}_m\right|_1+\lambda_m\left|\mathbf{C}_m-\hat{\mathbf{C}}_m\right|_1, \  \min_{\mathbf{R}, \mathbf{t}, \mathbf{S}}(\mathcal{L}_{ba}).
            \end{aligned}
        \end{math}
    }
    \label{eq:pose_grad}
\end{equation}


\section{Experiment}
\label{sec:experiment}
\subsection{Experimental Setup}
\noindent\textbf{Dataset.} To evaluate the performance of \ours, we conduct experiments on the Replica~\cite{straub2019replica}, and TUM-RGBD~\cite{Sturm2012ASystems}. Following~\cite{Zhu2021NICESLAMNI,Yang2022VoxFusionDT,Wang2023CoSLAMJC,Johari2022ESLAMED,pointslam}, we use 8 scenes from the Replica dataset for localization, mesh reconstruction, and rendering quality comparison. The selected three subsets of TUM-RGBD datasets are used for localization.

\noindent\textbf{Baselines.} We compare our method with existing SOTA NeRF-based dense visual SLAM: NICE-SLAM~\cite{Zhu2021NICESLAMNI}, Vox-Fusion~\cite{Yang2022VoxFusionDT}, CoSLAM~\cite{Wang2023CoSLAMJC}, ESLAM~\cite{Johari2022ESLAMED} and Point-SLAM~\cite{pointslam}. The rendering performance of CoSLAM~\cite{Wang2023CoSLAMJC} and ESLAM~\cite{Johari2022ESLAMED} is conducted from the open source code with the same configuration in \cite{pointslam}.

\noindent\textbf{Metric.} For mesh reconstruction, we use the 2D Depth L1 (cm)~\cite{Zhu2021NICESLAMNI}, the Precision (P, \%), Recall (R, \%), and F-score with a threshold of $1$ cm to measure the scene geometry. For localization, we use the absolute trajectory (ATE, cm) error~\cite{Sturm2012ASystems} to measure the accuracy of the estimated camera poses. We further evaluate the rendering performance using the peak signal-to-noise ratio (PSNR), SSIM~\cite{wang2004image}, and LPIPS~\cite{zhang2018unreasonable} by following \cite{pointslam}. To be fair, we run all the methods on a dataset 10 times and report the average results. More details can be found in the supplemental materials.

\noindent \textbf{Implementation details.} \ours is implemented in Python using the PyTorch framework, incorporating CUDA code for Gaussian splatting and trained on a desktop PC with a 5.50GHz Intel Core i9-13900K CPU and NVIDIA RTX 4090 GPU. We extended the existing code for differentiable Gaussian splatting rasterization with additional functionality for handling depth, pose, and cumulative opacity during both forward and backward propagation.  More details can be found in the supplemental materials.

\subsection{Evaluation of Localization and Mapping}
\label{sec:evaluation}
\boldparagraph{Evaluation on Replica.} \textbf{\emph{Tracking ATE}}: \cref{tab:tracking_replica} illustrates the tracking performance of our method and the state-of-the-art methods on the Replica dataset. Our method achieves the best or second performance in 7 of 8 scenes and outperforms the second-best method Point-SLAM~\cite{pointslam} by 0.4 cm on average at 8.34 FPS. It is noticeable that the second best method, Point-SLAM~\cite{pointslam} runs at 0.42 FPS, which is 20$\times$ slower than our method, indicating that \ours achieves a better trade-off between the tracking accuracy and the runtime efficiency. \textbf{\emph{Mapping ACC:}} \cref{tab:recon_replica} report the mapping evaluation results of our method with other current state-of-the-art visual SLAM methods. \ours achieves the best performance in Depth L1 (1.16cm) and Precision (74.0\%) metrics on average. For Recall and F1 scores, \ours performs comparably to the second best method CoSLAM~\cite{Wang2023CoSLAMJC}. The visualization results in~\cref{fig:mesh_replica} show that \ours achieves satisfying construction mesh with clear boundaries and details.

\boldparagraph{Evaluation on TUM-RGBD.} \cref{tab:tracking_tumrgbd} compares \ours with the other SLAM systems in TUM-RGBD dataset. Our method surpasses iMAP~\cite{Sucar2021iMAPIM}, NICE-SLAM~\cite{Zhu2021NICESLAMNI} and Vox-fusion~\cite{Yang2022VoxFusionDT}, and achieves a comparable performance, average 3.7 cm ATE RSME, with the SOTA methods. A gap to traditional methods still exists between the neural vSLAM and the traditional SLAM systems, which employ more sophisticated tracking schemes~\cite{pointslam}.

\begin{table}[h]
    \centering
    \vspace{-0ex}
    \caption{Tracking comparison (ATE RMSE [cm]) of the proposed method vs. the SOTA methods on the Replica dataset. The running speed of methods in the upper part is lower than 5 FPS, $^*$ denotes the reproduced results by running officially released code.}
    \label{tab:tracking_replica}
    \resizebox{0.48\textwidth}{!}{
        \begin{tabular}{r|ccccccccc}
            \hline
            Method                                    & \texttt{Rm0}      & \texttt{Rm1}      & \texttt{Rm2}      & \texttt{Off0}     & \texttt{Off1}     & \texttt{Off2}     & \texttt{Off3}     & \texttt{Off4}     & \texttt{\textbf{avg}} \\
            \hline
            Point-SLAM~\cite{pointslam}               & \nd 0.56          & \fs \textbf{0.47} & \fs \textbf{0.30} & \fs \textbf{0.35} & \rd 0.62          & \fs \textbf{0.55} & \nd 0.72          & \rd 0.73          & \nd 0.54              \\
            NICE-SLAM~\cite{Zhu2021NICESLAMNI}        & 0.97              & 1.31              & 1.07              & 0.88              & 1.00              & 1.06              & \rd 1.10          & 1.13              & 1.06                  \\
            Vox-Fusion$^*$~\cite{Yang2022VoxFusionDT} & 1.37              & 4.70              & 1.47              & 8.48              & 2.04              & 2.58              & 1.11              & 2.94              & 3.09                  \\
            \hline
            ESLAM~\cite{Johari2022ESLAMED}            & 0.71              & \rd 0.70          & \rd 0.52          & \rd 0.57          & \nd 0.55          & \nd 0.58          & \nd 0.72          & \fs \textbf{0.63} & \rd 0.63              \\
            CoSLAM~\cite{Wang2023CoSLAMJC}            & \rd 0.70          & 0.95              & 1.35              & 0.59              & \nd 0.55          & 2.03              & 1.56              & 0.72              & 1.00                  \\
            Ours                                      & \fs \textbf{0.48} & \nd 0.53          & \nd 0.33          & \nd 0.52          & \fs \textbf{0.41} & \rd 0.59          & \fs \textbf{0.46} & \nd 0.7           & \fs\textbf{0.50}      \\
            \hline
        \end{tabular}
    }
    \vspace{-1ex}
\end{table}

\begin{table}[h]
    \centering
    \vspace{-1ex}
    \caption{Tracking ATE [cm] on TUM-RGBD~\cite{Sturm2012ASystems}. Our method achieves a comparable performance among the neural vSLAMs. $^*$ denotes the reproduced results by running officially released code.}
    \label{tab:tracking_tumrgbd}
    \setlength{\tabcolsep}{2pt}
    \renewcommand{\arraystretch}{1}
    \begin{minipage}{.5\linewidth}
        \centering
        \resizebox{\columnwidth}{!}{
            \begin{tabular}{l|cccc}
                \toprule
                Method & \texttt{fr1\_desk} & \texttt{fr2\_xyz} & \texttt{fr3\_off} & Avg. \\
                \midrule
                DI-Fusion~\cite{huang2021di} & 4.4 & 2.0 & 5.8 & 4.1 \\
                ElasticFusion~\cite{whelan2015elasticfusion} & \rd{2.5} & \rd{1.2} & \rd{2.5} & \rd{2.1} \\
                BAD-SLAM~\cite{schops2019bad} & \nd{1.7} & \nd{1.1} & \nd{1.7} & \nd{1.5} \\
                Kintinuous~\cite{whelan2012kintinuous} & 3.7 & 2.9 & 3.0 & 3.2 \\
                ORB-SLAM2~\cite{MurArtal2016ORBSLAM2AO} & \fs{\textbf{1.6}} & \fs{\textbf{0.4}} & \fs{\textbf{1.0}} & \fs{\textbf{1.0}} \\
                \bottomrule
                iMAP$^*$~\cite{Sucar2021iMAPIM} & 7.2 & 2.1 & 9.0 & 6.1 \\
                \bottomrule
            \end{tabular}
            
        }
    \end{minipage}%
    \begin{minipage}{.5\linewidth}
        \centering
        \resizebox{\columnwidth}{!}{
            \begin{tabular}{l|cccc}
                \toprule
                Method & \texttt{fr1\_desk} & \texttt{fr2\_xyz} & \texttt{fr3\_off} & Avg. \\
                \midrule
                NICE-SLAM~\cite{Zhu2021NICESLAMNI} & 4.3 & 31.7 & 3.9 & 13.3 \\
                Vox-Fusion$^*$~\cite{Yang2022VoxFusionDT} & 3.5 & \rd{1.5} & 26.0 & 10.3 \\
                CoSLAM~\cite{Wang2023CoSLAMJC} & \nd{2.7} & 1.9 & \nd{2.6} & \nd{2.4} \\
                ESLAM~\cite{Johari2022ESLAMED} & \fs{2.3} & \fs{1.1} & \fs{2.4} & \fs{2.0} \\
                Point-SLAM & \nd{2.6} & \nd{1.3} & \rd{3.2} & \nd{2.4} \\
                \textbf{Ours} & \rd{3.3} & \nd{1.3} & 6.6 & \rd{3.7} \\
                \bottomrule
            \end{tabular}
        }
    \end{minipage}
    \vspace{-2ex}
\end{table}

\begin{table}[h]
    \vspace{2ex}
    \centering
    \caption{Reconstruction comparison of the proposed method vs. the SOTA methods on Replica dataset.}
    \label{tab:recon_replica}
    \resizebox{0.48\textwidth}{!}{
        \setlength{\tabcolsep}{2pt}
        \renewcommand{\arraystretch}{1}
        \begin{tabular}{l|lrrrrrrrrr}
            \hline
            Method                                                                                          & Metric                & \texttt{Rm\thinspace0} & \texttt{Rm\thinspace1} & \texttt{Rm\thinspace2} & \texttt{Off\thinspace0} & \texttt{Off\thinspace1} & \texttt{Off\thinspace2} & \texttt{Off\thinspace3} & \texttt{Off\thinspace4} & Avg.      \\
            \hline
            \multirow{4}{*}{\begin{tabular}[l]{@{}c@{}}NICESL\\AM~\cite{Zhu2021NICESLAMNI}\end{tabular}}    & Depth L1 $\downarrow$ & 1.81                   & \rd 1.44               & \rd 2.04               & 1.39                    & 1.76                    & 8.33                    & 4.99                    & 2.01                    & 2.97      \\
                                                                                                            & Precision $\uparrow$  & 45.86                  & 43.76                  & 44.38                  & 51.40                   & 50.80                   & 38.37                   & 40.85                   & 37.35                   & 44.10     \\
                                                                                                            & Recall$\uparrow$      & 44.10                  & 46.12                  & 42.78                  & 48.66                   & 53.08                   & 39.98                   & 39.04                   & 35.77                   & 43.69     \\
                                                                                                            & F1$\uparrow$          & 44.96                  & 44.84                  & 43.56                  & 49.99                   & 51.91                   & 39.16                   & 39.92                   & 36.54                   & 43.86     \\
            \hline
            \multirow{4}{*}{\begin{tabular}[l]{@{}c@{}}VoxFus\\ion~\cite{Yang2022VoxFusionDT}\end{tabular}} & Depth L1$\downarrow$  & \rd 1.09               & 1.90                   & 2.21                   & 2.32                    & 3.40                    & \nd 4.19                & \nd 2.96                & 1.61                    & \rd 2.46  \\
                                                                                                            & Precision$\uparrow$   & \nd 75.83              & 35.88                  & 63.10                  & 48.51                   & 43.50                   & 54.48                   & \nd 69.11               & 55.40                   & 55.73     \\
                                                                                                            & Recall$\uparrow$      & \rd 64.89              & 33.07                  & 56.62                  & 44.76                   & 38.44                   & 47.85                   & \rd 60.61               & 46.79                   & 49.13     \\
                                                                                                            & F1$\uparrow$          & \rd 69.93              & 34.38                  & 59.67                  & 46.54                   & 40.81                   & 50.95                   & \nd 64.56               & 50.72                   & 52.20     \\
            \hline
            \multirow{4}{*}{\begin{tabular}[l]{@{}c@{}}CoSLA\\M~\cite{Wang2023CoSLAMJC}\end{tabular}}       & Depth L1$\downarrow$  & \nd 0.99               & \nd 0.82               & 2.28                   & \rd 1.24                & \nd 1.61                & 7.70                    & 4.65                    & \rd 1.43                & 2.59      \\
                                                                                                            & Precision$\uparrow$   & \fs 81.71              & \nd 77.95              & \nd 73.30              & \nd 79.41               & \nd 80.67               & \rd 55.64               & 57.63                   & \fs 79.76               & \rd 73.26 \\
                                                                                                            & Recall$\uparrow$      & \nd 74.03              & \nd 70.79              & \nd 65.73              & \rd 71.46               & \rd 70.35               & \rd 52.96               & 56.06                   & \nd 71.22               & \rd 66.58 \\
                                                                                                            & F1$\uparrow$          & \nd 77.68              & \rd  74.20             & \nd 69.31              & \rd 75.23               & \nd 75.16               & \rd 54.27               & 56.83                   & \nd 75.25               & \rd 69.74 \\
            \hline
            \multirow{4}{*}{\begin{tabular}[l]{@{}c@{}}ESL\\AM~\cite{Johari2022ESLAMED}\end{tabular}}       & Depth L1$\downarrow$  & \fs 0.63               & \fs 0.62               & \fs 0.98               & \fs 0.57                & \rd 1.66                & \rd 7.32                & \rd 3.94                & \fs 0.88                & \nd 2.08  \\
                                                                                                            & Precision$\uparrow$   & \rd 74.33              & \rd 75.94              & \fs 82.48              & \rd 72.20               & \rd 65.74               & \nd 70.73               & \fs 72.48               & \nd 72.24               & \nd 73.27 \\
                                                                                                            & Recall$\uparrow$      & \fs 87.37              & \fs 87.01              & \fs 84.99              & \fs 88.36               & \fs 84.38               & \fs 81.92               & \fs 79.18               & \fs 80.63               & \fs 84.23 \\
                                                                                                            & F1$\uparrow$          & \fs 80.32              & \fs 81.10              & \fs 83.72              & \nd 79.47               & \rd 73.90               & \fs 75.92               & \fs 75.68               & \fs 76.21               & \fs 78.29 \\
            \hline
            \multirow{4}{*}{Ours}
                                                                                                            & Depth L1$\downarrow$  & 1.31                   & \nd 0.82               & \nd 1.26               & \nd 0.81                & \fs 0.96                & \fs 1.41                & \fs 1.53                & \nd 1.08                & \fs 1.16  \\
                                                                                                            & Precision$\uparrow$   & 64.58                  & \fs 83.11              & \rd 70.13              & \fs 83.43               & \fs 87.77               & \fs 70.91               & \rd 63.18               & \rd 68.88               & \fs 74.00 \\
                                                                                                            & Recall$\uparrow$      & 61.29                  & \rd 76.83              & \rd 63.84              & \nd 76.90               & \nd 76.15               & \nd 61.63               & \nd 62.91               & \rd 61.50               & \nd 67.63 \\
                                                                                                            & F1$\uparrow$          & 62.89                  & \nd 79.85              & \rd 66.84              & \fs 80.03               & \fs 81.55               & \nd 65.95               & \rd 59.17               & \rd 64.98               & \nd 70.15 \\
            \hline
        \end{tabular}
    }
    \vspace{-3ex}
\end{table}

\subsection{Rendering Evaluation}
\begin{figure*}[h]
    \begin{center}
        \includegraphics[width=1\linewidth]{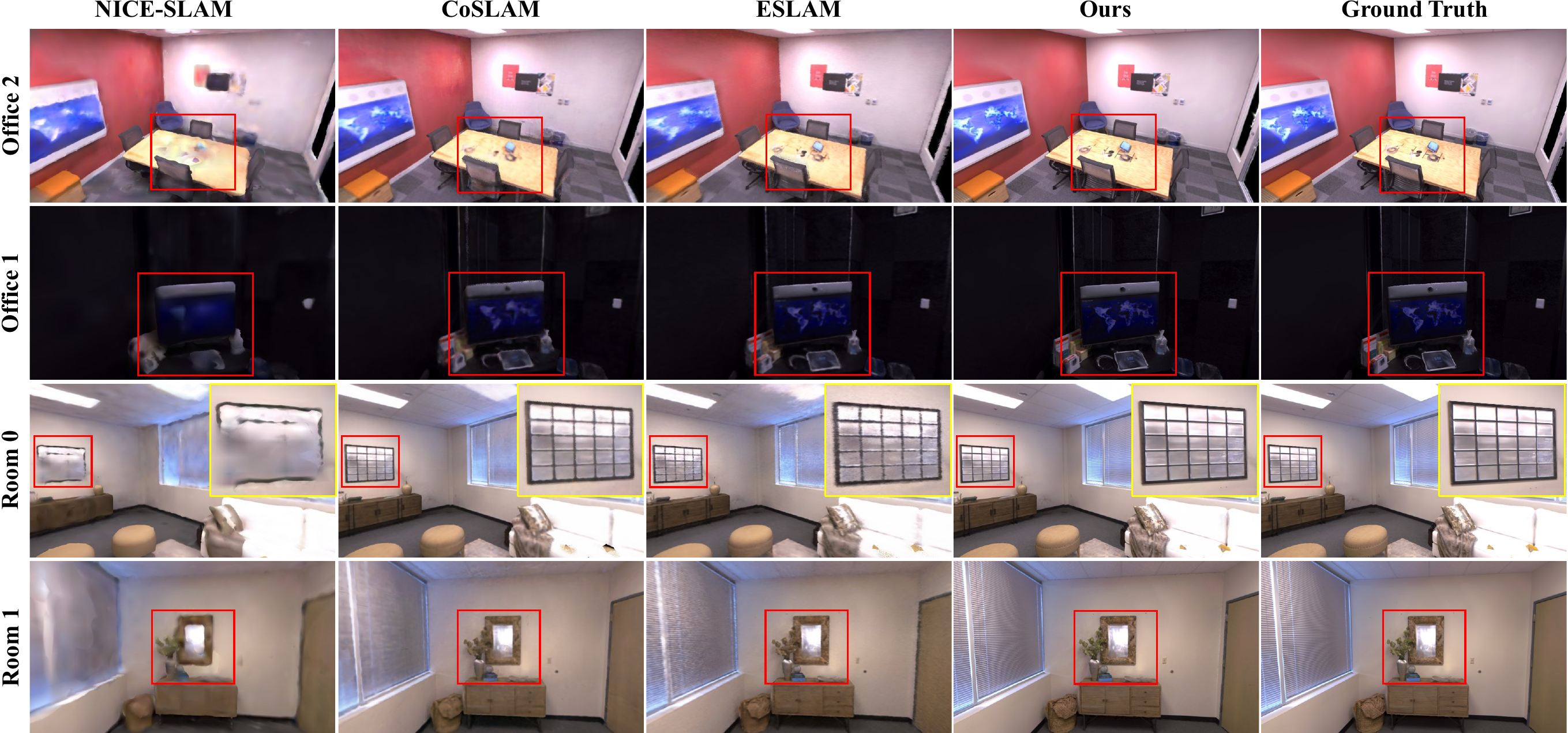}
    \end{center}
    \vspace{-2ex}
    \caption{The render visualization results on the Replica dataset of the proposed \ours and SOTA methods. \ours can generate much more high-quality and realistic images than the other methods, especially around the object boundaries.}
    \label{fig:render_replica}
    \vspace{-2ex}
\end{figure*}

We compare the rendering performance of the proposed \ours with the neural visual SLAM methods in~\cref{tab:rendering_replica}. The results show that \ours achieves the best performance in all the metrics. Our method significantly outperforms the second-best methods CoSLAM~\cite{Wang2023CoSLAMJC}, ESLAM~\cite{Johari2022ESLAMED} and NICE-SLAM~\cite{Zhu2021NICESLAMNI} by 1.52 dB in PSNR, 0.027 in SSIM and 0.12 in LPIPS, respectively. It is noticeable that \ours achieves 386 FPS rendering speed on average, which is 100$\times$ faster than the second-best method Vox-Fusion~\cite{Yang2022VoxFusionDT}. This excellent rendering performance is attributed to the efficient 3D Gaussian rendering pipeline and can be further applied to real-time downstream tasks, such as VR~\cite{Desai2014ARP}, robot navigation~\cite{Hne2011StereoDM} and autonomous driving~\cite{Bresson2017SimultaneousLA}. The visualization results in~\cref{fig:render_replica} show that \ours can generate much more high-quality and realistic images than the other methods, especially in edge areas with detailed structures. While NICE-SLAM~\cite{Zhu2021NICESLAMNI} causes severe artifacts and blurs, CoSLAM~\cite{Wang2023CoSLAMJC} and ESLAM~\cite{Johari2022ESLAMED} generate blur around the image boundaries.

\begin{figure}[h]
    \vspace{-2ex}
    \centering
    {\footnotesize
        \setlength{\tabcolsep}{0.25pt}
        \renewcommand{\arraystretch}{0}
        \newcommand{\sz}{0.3}
        \resizebox{1\columnwidth}{!}{
            \begin{tabular}{ccccc}
                                                                                        & \texttt{Room 0}                                                                  & \texttt{Room 1}                                                                  & \texttt{Room 2}                                                                  & \texttt{Office 3}                                                                  \\
                \rotatebox[origin=c]{90}{{\tiny NICE-SLAM~\cite{Zhu2021NICESLAMNI}}}    & \includegraphics[valign=c,width=\sz\linewidth]{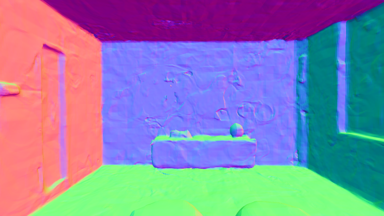}  & \includegraphics[valign=c,width=\sz\linewidth]{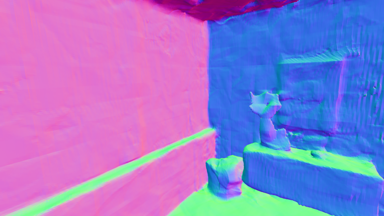}  & \includegraphics[valign=c,width=\sz\linewidth]{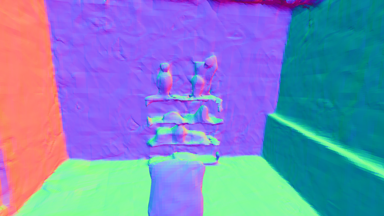}  & \includegraphics[valign=c,width=\sz\linewidth]{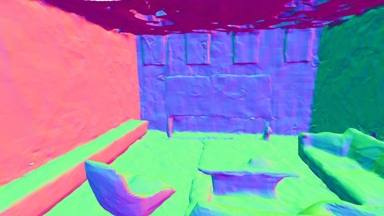}  \\
                \rotatebox[origin=c]{90}{{\tiny Vox-Fusion~\cite{Yang2022VoxFusionDT}}} & \includegraphics[valign=c,width=\sz\linewidth]{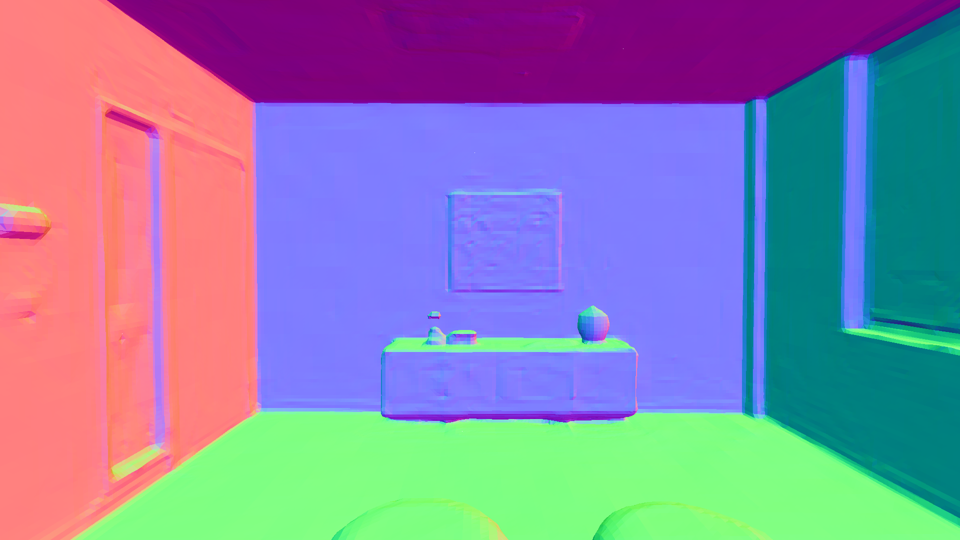} & \includegraphics[valign=c,width=\sz\linewidth]{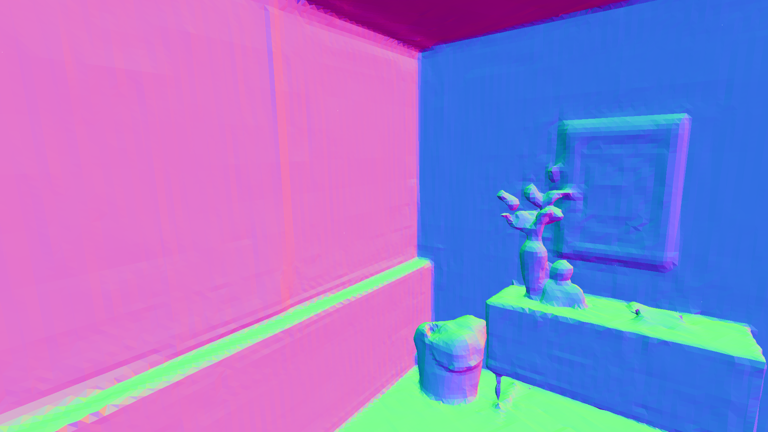} & \includegraphics[valign=c,width=\sz\linewidth]{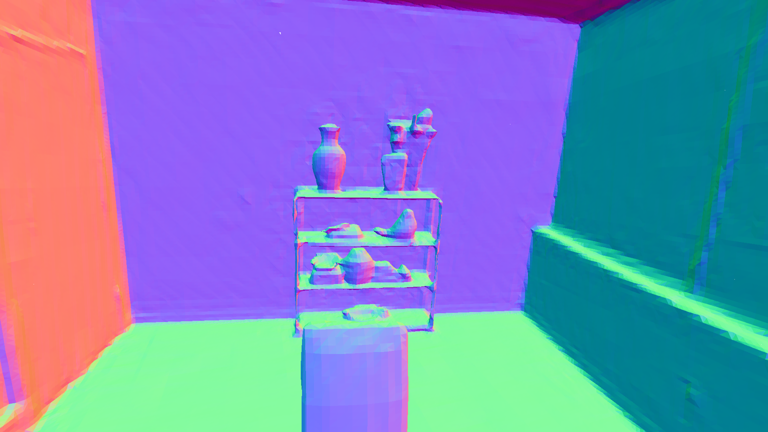} & \includegraphics[valign=c,width=\sz\linewidth]{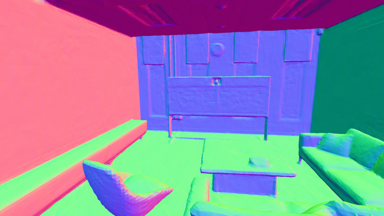} \\
                \rotatebox[origin=c]{90}{{\tiny CoSLAM~\cite{Wang2023CoSLAMJC}}}        & \includegraphics[valign=c,width=\sz\linewidth]{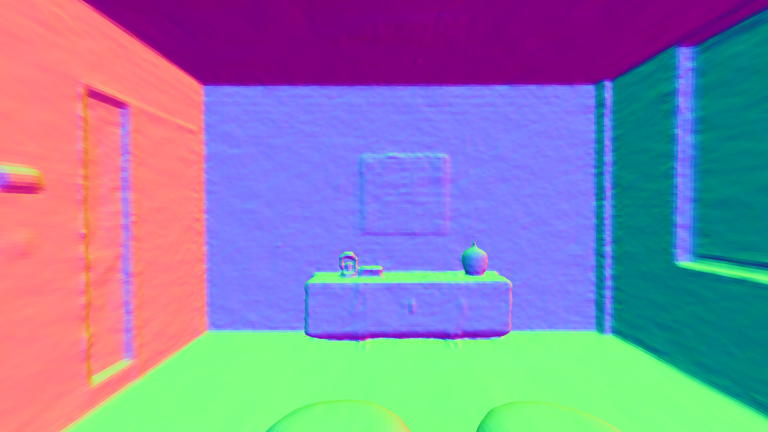}    & \includegraphics[valign=c,width=\sz\linewidth]{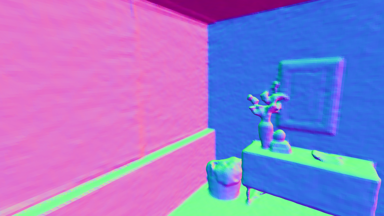}    & \includegraphics[valign=c,width=\sz\linewidth]{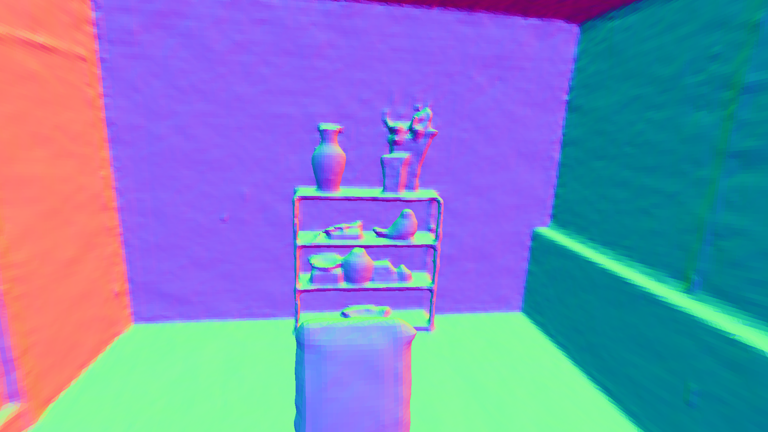}    & \includegraphics[valign=c,width=\sz\linewidth]{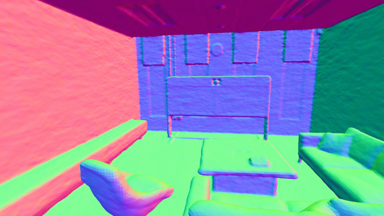}    \\                                                                                                                                                                                                                                                                                                                                                                            \\
                \rotatebox[origin=c]{90}{{\tiny ESLAM~\cite{Johari2022ESLAMED}}}        & \includegraphics[valign=c,width=\sz\linewidth]{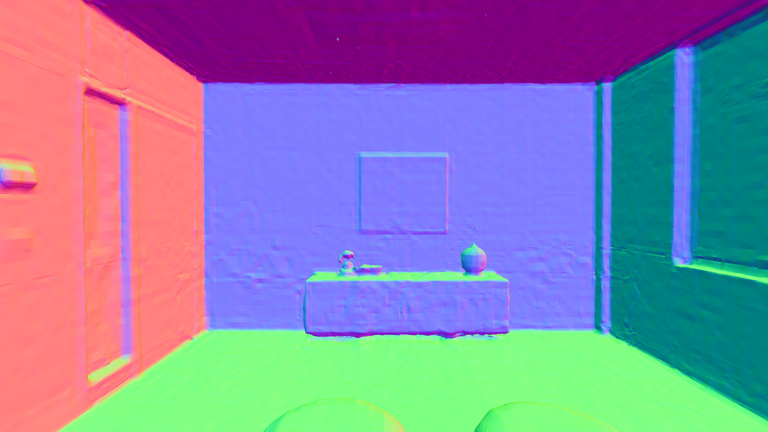}     & \includegraphics[valign=c,width=\sz\linewidth]{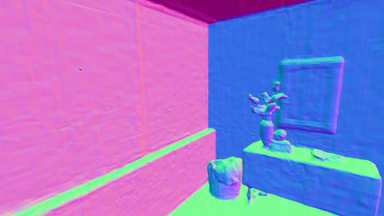}     & \includegraphics[valign=c,width=\sz\linewidth]{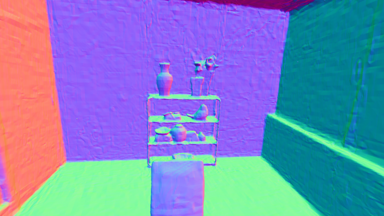}     & \includegraphics[valign=c,width=\sz\linewidth]{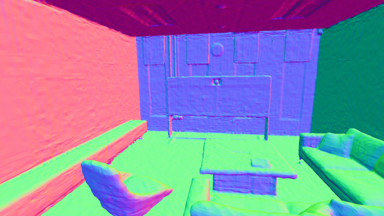}     \\                                                                                                                                                                                                                                                                                                                                                                            \\
                \rotatebox[origin=c]{90}{{\tiny Ours}}                                  & \includegraphics[valign=c,width=\sz\linewidth]{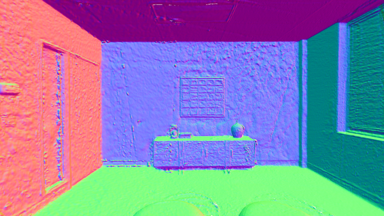}    & \includegraphics[valign=c,width=\sz\linewidth]{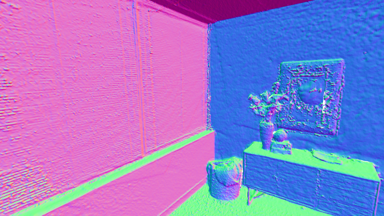}    & \includegraphics[valign=c,width=\sz\linewidth]{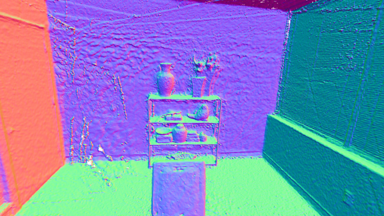}    & \includegraphics[valign=c,width=\sz\linewidth]{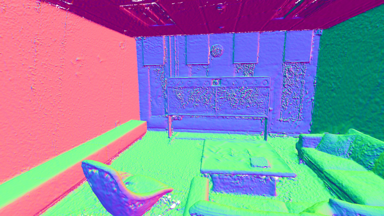}    \\                                                                                                                                                                                                                                                                                                                                                                            \\
                \rotatebox[origin=c]{90}{{\tiny Ground Truth}}                          & \includegraphics[valign=c,width=\sz\linewidth]{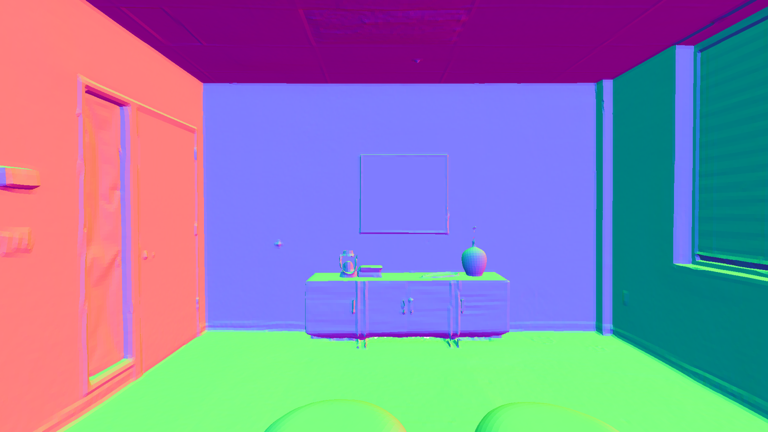}        & \includegraphics[valign=c,width=\sz\linewidth]{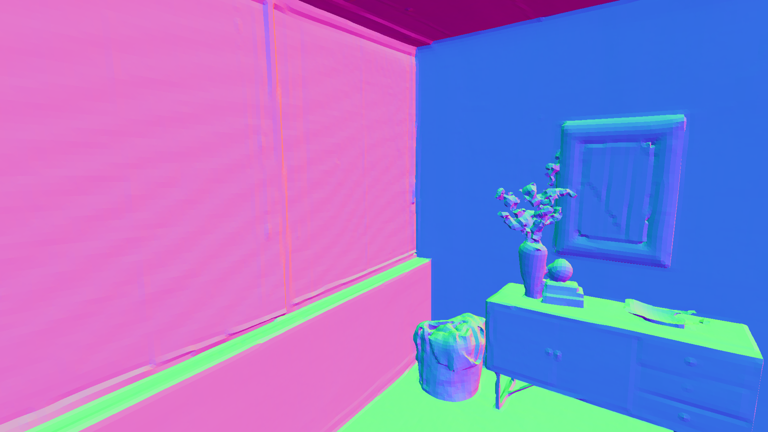}        & \includegraphics[valign=c,width=\sz\linewidth]{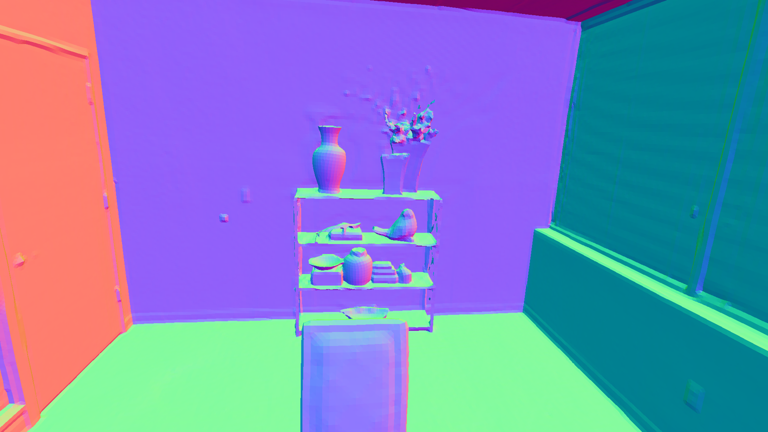}        & \includegraphics[valign=c,width=\sz\linewidth]{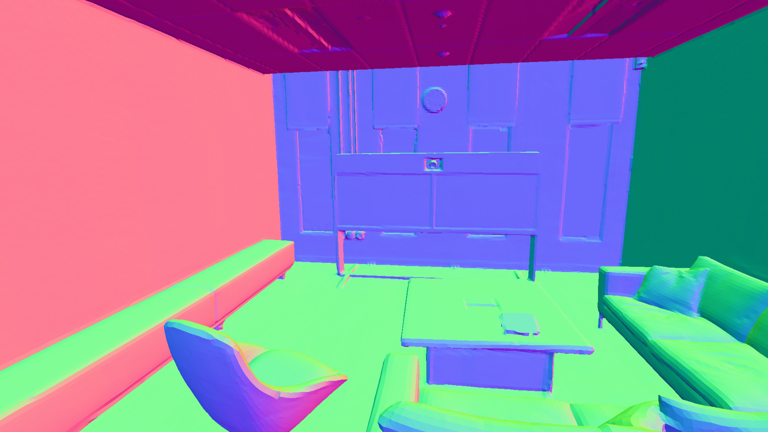}        \\                                                                                                                                                                                                                                                                                                                                                                            \\
            \end{tabular}
        }
    }
    \caption{Reconstruction performance comparation of the proposed \ours and SOTA methods on the Replica dataset.}
    \label{fig:mesh_replica}
    \vspace{-3ex}
\end{figure}

\label{sec:render}

\subsection{Runtime Analysis} \cref{tab:memory_runtime} and \cref{tab:rebuttal_memory_usage} illustrate the runtime and memory usage of \ours and the state-of-the-art methods on the Replica and TUM-RGBD, respectively. We report the parameters of the neural networks and the memory usage of the scene representation. Note that Point-SLAM uses extra memory dynamic radius to improve performance (mark as $^\dagger$). The results show that \ours achieves a competitive running speed with 8.34 FPS compared to the other Radiance Fields-based vSLAMs. Note that we do not use any neural network decoder in our system, which results in the zero learnable parameter. However, the 3D Gaussian scene representations of \ours consume 198.04 MB memory, 4$\times$ larger than the second large method NICE-SLAM~\cite{Zhu2021NICESLAMNI}. Memory usage is mainly caused by spherical harmonic coefficients in training, which is a common constraint among Gaussian splatting-based reconstruction methods. Despite this, we still achieve a 20 $\times$ faster FPS compared to the similar point-based method Point-SLAM~\cite{pointslam}. Besides, we also provide a light version of \ours with zero-order spherical harmonic coefficients, significantly reducing memory usage while maintaining stable performance.

\begin{table}[t]
    \vspace{2ex}
    \centering
    \caption{Runtime and memory usage on Replica \texttt{\#Room0}. The decoder parameters and embedding denote the parameter number of MLPs and the memory usage of the scene representation. }
    \scriptsize
    \setlength{\tabcolsep}{5pt}
    \resizebox{\columnwidth}{!}{
        \begin{tabular}{lcccccccc}
            \toprule
            \multirow{2}{*}{Method}               & Tracking                    & Mapping                     & System         & Decoder            & Scene                 \\
                                                  & [ms$\times$it] $\downarrow$ & [ms$\times$it] $\downarrow$ & FPS~$\uparrow$ & param $\downarrow$ & Embedding$\downarrow$ \\
            \midrule
            Point-SLAM~\cite{pointslam}           & 0.06 $\times$ 40            & 34.81 $\times$ 300          & 0.42           & 0.127 M            & 55.42 (+12453.2)$^\dagger$MB                \\
            NICE-SLAM~\cite{Zhu2021NICESLAMNI}    & 6.64 $\times$ 10            & 28.63 $\times$ 60           & 2.91           & 0.06 M             & 48.48 MB              \\
            Vox-Fusion~\cite{Yang2022VoxFusionDT} & 0.03 $\times$ 30            & 66.53 $\times$ 10           & 1.28           & 0.054 M            & 1.49 MB               \\
            \midrule
            CoSLAM~\cite{Zhu2021NICESLAMNI}       & 6.01 $\times$ 10            & 13.18 $\times$ 10           & 16.64          & 1.671 M            & ---                   \\
            ESLAM~\cite{Johari2022ESLAMED}        & 6.85 $\times$ 8             & 19.87 $\times$ 15           & 13.42          & 0.003 M            & 27.12 MB              \\
            \ours                                 & 11.9 $\times$ 10            & 12.8 $\times$ 100           & 8.34           & \textbf{0 M}       & 198.04 MB             \\
            \bottomrule
        \end{tabular}
    }
    \label{tab:memory_runtime}
\end{table}

\begin{table}[h]
    \vspace{-1.5ex}
    \centering
    \caption{Runtime and memory usage on TUM-RGBD dataset \texttt{\#fr1\_desk} and \texttt{\#fr2\_xyz}.}
    \scriptsize
    \setlength{\tabcolsep}{5pt}
    \resizebox{\columnwidth}{!}{
        \begin{tabular}{lcccccc}
            \toprule
            \multirow{2}{*}{Method}            & \multicolumn{2}{c}{\texttt{\#fr1\_desk}} &                                             & \multicolumn{2}{c}{\texttt{\#fr2\_xyz}} &                                                                                      \\ \cline{2-3} \cline{5-6}
                                               & \multirow{2}{*}[0.5ex]{FPS~$\uparrow$}   & \multirow{2}{*}[0.5ex]{Memory~$\downarrow$} &                                         & \multirow{2}{*}[0.5ex]{FPS~$\uparrow$} & \multirow{2}{*}[0.5ex]{Memory~$\downarrow$} \\
            \midrule
            Point-SLAM                         & 0.10                                     & 18.3 (+160.7)$^\dagger$MB                   &                                         & 0.12                                   & 14.2 (+7687.4)$^\dagger$MB                  \\
            NICE-SLAM~\cite{Zhu2021NICESLAMNI} & 0.11                                     & 178.8MB                                     &                                         & 0.12                                   & 484.0MB                                     \\
            ESLAM~\cite{Johari2022ESLAMED}     & 0.31                                     & 27.2MB                                      &                                         & 0.31                                   & 51.6MB                                      \\
            \midrule
            \ours                              & 1.83(ATE:3.3)                            & 40.8MB                                      &                                         & 1.51(ATE:1.3)                          & 48.4MB                                      \\
            \ours(light)                       & 1.92(ATE:4.3)                            & 18.8MB                                      &                                         & 1.68(ATE:2.7)                          & 22.3MB                                      \\
            \bottomrule
        \end{tabular}
    }
    \vspace{-6ex}
    \label{tab:rebuttal_memory_usage}
\end{table}

\begin{table*}[h]
    \vspace{-2ex}
    \centering
    \footnotesize
    \setlength{\tabcolsep}{2.65pt}
    \renewcommand{\arraystretch}{1}
    \caption{Rendering performance on Replica dataset. We outperform existing dense neural RGB-D methods on the commonly reported rendering metrics. Note that \ours achieves 386 FPS on average, benefiting from the efficient Gaussian scene representation.}
    \begin{tabularx}{\linewidth}{l|lcccccccccc}
        \hline
        Method                                                       & Metric               & \texttt{Room 0} & \texttt{Room 1} & \texttt{Room 2} & \texttt{Office 0} & \texttt{Office 1} & \texttt{Office 2} & \texttt{Office 3} & \texttt{Office 4} & \texttt{Avg.} & \texttt{FPS.} \\
        \hline
        \multirow{3}{*}{NICE-SLAM~\cite{Zhu2021NICESLAMNI}}
                                                                     & PSNR [dB] $\uparrow$ & 22.12           & 22.47           & 24.52           & 29.07             & \rd 30.34         & 19.66             & 22.23             & 24.94             & 24.42         &               \\
                                                                     & SSIM $\uparrow$      & 0.689           & 0.757           & \rd 0.814       & 0.874             & 0.886             & 0.797             & 0.801             & 0.856             & 0.809         & 0.30          \\
                                                                     & LPIPS $\downarrow$   & 0.330           & 0.271           & \nd 0.208       & 0.229             & \nd 0.181         & \nd 0.235         & \rd 0.209         & \nd 0.198         & \nd 0.233     &               \\
        \hline
        \multirow{3}{*}{Vox-Fusion$^{*}$~\cite{Yang2022VoxFusionDT}} & PSNR [dB] $\uparrow$ & 22.39           & 22.36           & 23.92           & 27.79             & 29.83             & 20.33             & 23.47             & 25.21             & 24.41         & \nd           \\
                                                                     & SSIM $\uparrow$      & 0.683           & 0.751           & 0.798           & 0.857             & 0.876             & 0.794             & 0.803             & 0.847             & 0.801         & \nd 3.88      \\
                                                                     & LPIPS $\downarrow$   & \nd 0.303       & \nd 0.269       & \rd 0.234       & 0.241             & \rd 0.184         & 0.243             & 0.213             & \rd 0.199         & \rd 0.236     & \nd           \\
        \hline
        \multirow{3}{*}{CoSLAM~\cite{Wang2023CoSLAMJC}}
                                                                     & PSNR [dB] $\uparrow$ & \nd 27.27       & \nd 28.45       & \rd 29.06       & \nd 34.14         & \nd 34.87         & \nd 28.43         & \rd 28.76         & \nd 30.91         & \nd 30.24     & \rd           \\
                                                                     & SSIM $\uparrow$      & \nd 0.910       & \nd 0.909       & \nd 0.932       & \nd 0.961         & \nd 0.969         & \rd 0.938         & \rd 0.941         & \nd 0.955         & \nd 0.939     & \rd 3.68      \\
                                                                     & LPIPS $\downarrow$   & 0.324           & 0.294           & 0.266           & \rd 0.209         & 0.196             & 0.258             & 0.229             & 0.236             & 0.252         & \rd           \\
        \hline
        \multirow{3}{*}{ESLAM~\cite{Johari2022ESLAMED}}
                                                                     & PSNR [dB] $\uparrow$ & \rd 25.32       & \rd 27.77       & \nd 29.08       & \rd 33.71         & 30.20             & \rd 28.09         & \nd 28.77         & \rd 29.71         & \rd 29.08     &               \\
                                                                     & SSIM $\uparrow$      & \rd 0.875       & \rd 0.902       & \nd 0.932       & \rd 0.960         & \rd 0.923         & \nd 0.943         & \nd 0.948         & \rd 0.945         & \rd 0.929     & 2.82          \\
                                                                     & LPIPS $\downarrow$   & \rd 0.313       & \rd 0.298       & 0.248           & \nd 0.184         & 0.228             & \rd 0.241         & \nd 0.196         & 0.204             & 0.336         &               \\

        \hline

        \multirow{3}{*}{Ours}
                                                                     & PSNR [dB] $\uparrow$ & \fs 31.56       & \fs 32.86       & \fs 32.59       & \fs 38.70         & \fs 41.17         & \fs 32.36         & \fs 32.03         & \fs 32.92         & \fs 34.27     & \fs           \\
                                                                     & SSIM $\uparrow$      & \fs 0.968       & \fs 0.973       & \fs 0.971       & \fs 0.986         & \fs 0.993         & \fs 0.978         & \fs 0.970         & \fs 0.968         & \fs 0.975     & \fs386.91     \\
                                                                     & LPIPS $\downarrow$   & \fs 0.094       & \fs 0.075       & \fs 0.093       & \fs 0.050         & \fs 0.033         & \fs 0.094         & \fs 0.110         & \fs 0.112         & \fs 0.082     & \fs           \\

        \hline
    \end{tabularx}
    \vspace{-3ex}
    \label{tab:rendering_replica}
\end{table*}

\subsection{Ablation Study}
\label{sec:ablation}

We perform the ablation of \ours on the Replica dataset \texttt{\#Room0} subset to evaluate the effectiveness of coarse-to-fine tracking, and expansion mapping strategy.

\begin{figure}[h]
    \vspace{-2ex}
    \begin{center}
        \includegraphics[width=0.9\linewidth]{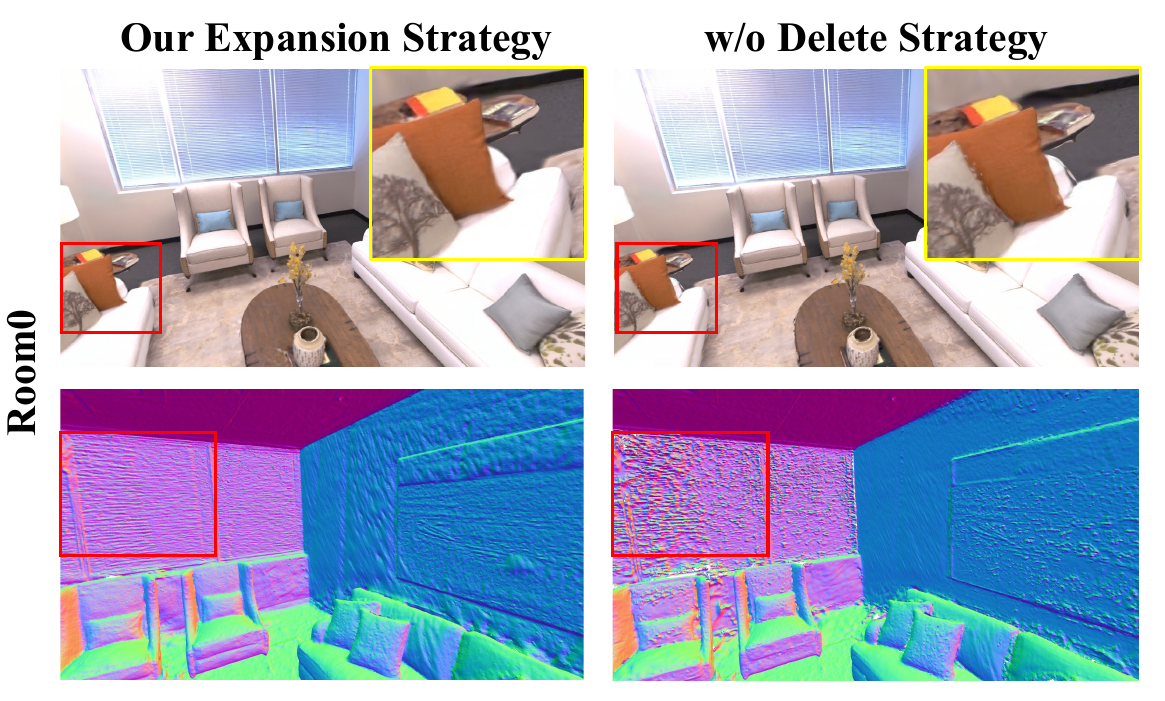}
    \end{center}
    \vspace{-3ex}
    \caption{Rendering and mesh visualization of the adaptive 3D Gaussian expansion ablation on Replica \texttt{\#Room0}.}
    \label{fig:ba_delete}
\end{figure}

\noindent{\textbf{Effect of our expansion strategy for mapping.}}
\cref{tab:ab_mapping} shows the ablation of our proposed expansion strategy for mapping. The results illustrate that the expansion strategy can significantly improve the tracking and mapping performance. The implementation w/o adding means that we only initialize 3D Gaussians in the first frame and optimize the scene without adding new points. However, this strategy completely crashes because the density control in~\cite{kerbl3Dgaussians} can not handle real-time mapping tasks without an accurate point cloud input. Besides, the implementation w/o deletion suffers from a large number of redundant and noisy 3D Gaussian, which causes undesirable supervision. In contrast, the proposed expansion strategy effectively improves the tracking and mapping performance by 0.1 in ATE and 11.97 in Recall by adding more accurate constraints for the optimization. According to the visualization results in~\cref{fig:ba_delete}, our full implementation achieves more high-quality and detailed rendering and reconstruction results than the w/o delete strategy.

\noindent{\textbf{Effect of coarse-to-fine tracking.}}
According to the results in~\cref{tab:ab_tracking}, the proposed coarse-to-fine tracking strategy performs best in all tracking, mapping, and rendering metrics. Compared with fine tracking, the coarse-to-fine tracking strategy significantly improves the performance by 0.01 in tracking ATE, 2.11 in Recall, and 0.72 in PSNR. Although the fine strategy surpasses the coarse strategy in precision, it suffers from the artifacts and noise in the reconstructed scene, leading to a fluctuation optimization. The coarse-to-fine strategy effectively avoids noise reconstruction and improves accuracy and robustness.

\begin{table}[t]
    \centering
    \scriptsize
    \caption{Ablation of the adaptive 3D Gaussian expansion strategy on Replica \texttt{\#Room0}.}
    \vspace{-1ex}
    \resizebox{0.45\textwidth}{!}{
        \setlength{\tabcolsep}{2.5pt}
        \begin{tabular}{lcccccccc}
            \toprule
            \multirow{2}{*}{Setting} & \multicolumn{8}{c}{\texttt{\#Room0}}                                                                                                                                                                                                                                                                                                                  \\\cline{2-9}
                                     & \multirow{2}{*}[0.5ex]{ATE$\downarrow$} & \multirow{2}{*}[0.5ex]{Depth L1$\downarrow$} & \multirow{2}{*}[0.5ex]{Precision$\uparrow$} & \multirow{2}{*}[0.5ex]{Recall $\uparrow$} & \multirow{2}{*}[0.5ex]{F1$\uparrow$} & \multirow{2}{*}[0.5ex]{PSNR$\uparrow$} & \multirow{2}{*}[0.5ex]{SSIM$\uparrow$} & \multirow{2}{*}[0.5ex]{LPIPS$\downarrow$} \\
            \midrule
            w/o add                  & \ding{55}                               & \ding{55}                                    & \ding{55}                                   & \ding{55}                                 & \ding{55}                            & \ding{55}                              & \ding{55}                              & \ding{55}                                 \\
            w/o delete               & 0.58                                    & 1.68                                         & 53.55                                       & 49.32                                     & 51.35                                & 31.22                                  & 0.967                                  & 0.094                                     \\
            \midrule
            w/ add \& delete         & \textbf{0.48}                           & \textbf{1.31}                                & \textbf{64.58}                              & \textbf{61.29}                            & \textbf{62.89}                       & \textbf{31.56}                         & \textbf{0.968}                         & \textbf{0.094}                            \\
            \bottomrule
        \end{tabular}
    }
    \vspace{-1ex}
    \label{tab:ab_mapping}
\end{table}

\begin{table}[t]
    \centering
    \scriptsize
    \caption{Ablation of the coarse-to-fine tracking strategy on Replica \texttt{\#Room0}.}
    \vspace{-1ex}
    \resizebox{0.45\textwidth}{!}{
        \setlength{\tabcolsep}{2.5pt}
        \begin{tabular}{lcccccccc}
            \toprule
            \multirow{2}{*}{Setting} & \multicolumn{8}{c}{\texttt{\#Room0}}                                                                                                                                                                                                                                                                                                                  \\\cline{2-9}
                                     & \multirow{2}{*}[0.5ex]{ATE$\downarrow$} & \multirow{2}{*}[0.5ex]{Depth L1$\downarrow$} & \multirow{2}{*}[0.5ex]{Precision$\uparrow$} & \multirow{2}{*}[0.5ex]{Recall $\uparrow$} & \multirow{2}{*}[0.5ex]{F1$\uparrow$} & \multirow{2}{*}[0.5ex]{PSNR$\uparrow$} & \multirow{2}{*}[0.5ex]{SSIM$\uparrow$} & \multirow{2}{*}[0.5ex]{LPIPS$\downarrow$} \\
            \midrule
            Coarse                   & 0.91                                    & 1.48                                         & 59.68                                       & 57.54                                     & 56.50                                & 29.13                                  & 0.954                                  & 0.120                                     \\
            Fine                     & \nd 0.49                                & \nd 1.39                                     & \nd 62.61                                   & \nd 59.18                                 & \nd 61.29                            & \nd 30.84                              & \nd 0.964                              & \nd 0.096                                 \\
            Coarse-to-fine           & \fs 0.48                                & \fs 1.31                                     & \fs 64.58                                   & \fs 61.29                                 & \fs 62.89                            & \fs 31.56                              & \fs 0.968                              & \fs 0.094                                 \\
            \bottomrule
        \end{tabular}
    }
    \vspace{-4.5ex}
    \label{tab:ab_tracking}
\end{table}

\subsection{Efficiency-to-accuracy trade-off.}

\begin{figure}[htbp]
    \centering
    \begin{subfigure}[b]{0.45\linewidth}
        \includegraphics[width=\linewidth]{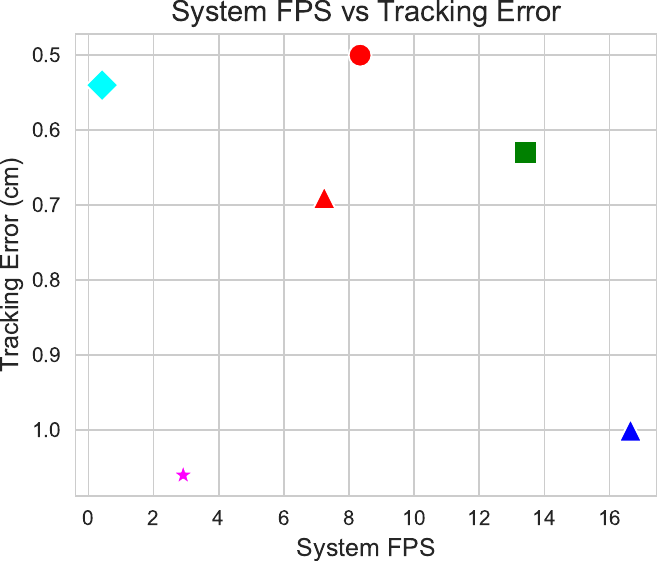}
        \vspace{-3ex}
        \caption{Tracking performance}
        \label{fig:bi_a}
    \end{subfigure}
    \hfill
    \begin{subfigure}[b]{0.45\linewidth}
        \includegraphics[width=\linewidth]{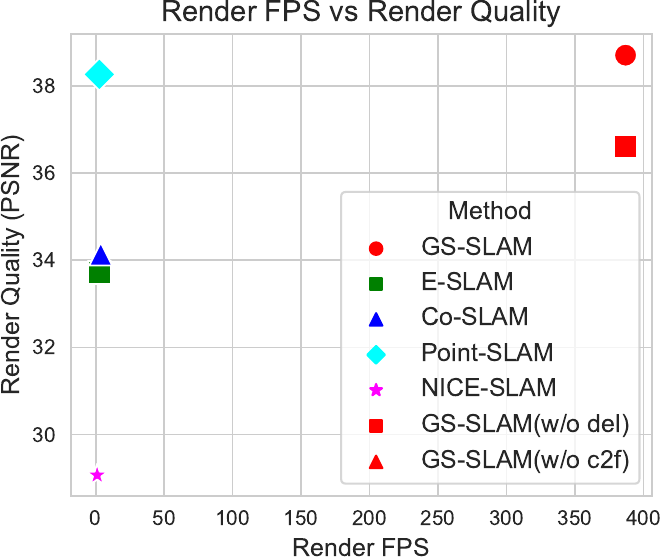}
        \vspace{-3ex}
        \caption{Render performance}
        \label{fig:bi_b}
    \end{subfigure}
    \vspace{1ex}
    \caption{Bi-criteria figure of tracking/render performance and system FPS on Replica \texttt{\#Office0}.}
    \label{fig:bi}
    \vspace{-2.5ex}
\end{figure}

3DGS-based SLAM trade-off focuses not only on the efficiency of tracking and mapping but also emphasizes high-quality ultra-real-time rendering. As shown in~\cref{fig:bi_b}, \ours achieves $\approx$ 400 FPS ultra-fast speed and highest PSNR in map rendering. At the same time, our method remains a competitive system FPS and lowest tracking ATE in~\cref{fig:bi_a}. Moreover, \ours shows great potential in memory reduction in~\cref{tab:rebuttal_memory_usage}, and comparable in mesh reconstruction. Note that baselines directly use 3DGS in~\cref{fig:bi}, resulting in inferior performances.
 
\section{Conclusion and Limitations}
\label{sec:limitation_conclusion}

We introduced \ours, a novel dense visual SLAM method leveraging 3D Gaussian Splatting for efficient mapping and accurate camera pose estimation, striking a better speed-accuracy balance.  However, its reliance on high-quality depth data may limit performance in certain conditions.  Additionally, the approach's high memory requirements for large scenes suggest future improvements could focus on optimizing memory use, potentially via techniques such as quantization and clustering. We believe GS-SLAM has the potential to extend to larger scale with some improvements and will explore this in future work.

{\small
\boldparagraph{Acknowledgements.} This work is supported by the Shanghai AI Laboratory, National Key R\&D Program of China (2022ZD0160101), the National Natural Science Foundation of China (62376222), Young Elite Scientists Sponsorship Program by CAST (2023QNRC001) and the Early Career Scheme of the Research Grants Council (RGC) of the Hong Kong SAR under grant No. 26202321.
}

\clearpage
\setcounter{page}{1}
\maketitlesupplementary

\setcounter{section}{0}
\begin{overview}
    This supplementary material complements our primary study, offering extended details and data to enhance the reproducibility of our GS-SLAM.  It also includes supplementary evaluations and a range of qualitative findings that further support the conclusions presented in the main paper:\\
    \noindent $\triangleright$ \textbf{\cref{sec:grad}}: Proofs of gradient computation, including the details of pose gradient derivation and depth gradient derivation.\\
    \noindent $\triangleright$ \textbf{\cref{sec:c2f_pose}}: Coarse-to-fine pose optimization algorithm, including pseudocode and visualization of how our algorithm worked.\\
    \noindent $\triangleright$ \textbf{\cref{sec:performance}}: Additional performance comparison, including the additional visualization on the Replica and TUM-RGBD dataset.\\
    \noindent $\triangleright$ \textbf{\cref{sec:full_ab}}: Additional ablation result, including the addition qualitative result on render, tracking and mapping performance.\\
    \noindent $\triangleright$ \textbf{\cref{sec:add_implement_detail}}: Additional implementation details, including more details for reproducibility of our GS-SLAM.\\
\end{overview}


\section{Proofs of Gradient Computation}
\label{sec:grad}
To derive gradients for pose $\mathbf{P} = \{ \mathbf{R}, \mathbf{t}\}$, recall that we store the pose in separate quaternion $\mathbf{q} \in \mathbb{R}^{4}$ and translation $\mathbf{t}\in \mathbb{R}^{3}$ vectors.
To derive gradients for rotation, we recall the conversion from a unit quaternion $\mathbf{q} = [q_r, q_i, q_j, q_k]^{T}$ to rotation matrix $\mathbf{R}$:
\begin{equation}\label{eq:q2R}
    \resizebox{0.87\linewidth}{!}{
        \begin{math}
            \begin{aligned}
                \mathbf{R} = 2\begin{bmatrix}
                                  \frac{1}{2} - (q_j^2 + q_k^2) & (q_i q_j - q_r q_k)           & (q_i q_k + q_r q_j)           \\
                                  (q_i q_j + q_r q_k)           & \frac{1}{2} - (q_i^2 + q_k^2) & (q_j q_k - q_r q_i)           \\
                                  (q_i q_k - q_r q_j)           & (q_j q_k + q_r q_i)           & \frac{1}{2} - (q_i^2 + q_j^2)
                              \end{bmatrix}
            \end{aligned}
        \end{math}
    }
\end{equation}

Also, we recall that the pose $\mathbf{P}$ is consisits of two part $\frac{\partial \mathbf{m}_i}{\partial {\mathbf{P}}}$ and $\frac{\partial\mathbf{\Sigma^{\prime}}}{\partial {\mathbf{P}}}$ when ignore view-dependent color. For simplicity of the formula, we denote $\mathbf{X}_i = [x, y, z]^{T}$ in the camera coordinate as $\mathbf{X}_i^c = \mathbf{P}\mathbf{X}_i = [x^c, y^c, z^c]^{T}$.

\vspace{0.5ex}
\boldparagraph{Pose gradients back-propagation by $\frac{\partial\mathbf{m}_i}{\partial{\mathbf{P}}}$.} We find the following gradients for translation $\mathbf{t}$ and $\frac{\partial \mathbf{X}_i^c}{\partial t}$:
\begin{equation}
    \begin{aligned}
         & \resizebox{0.60\linewidth}{!}{
            \begin{math}
                \frac{\partial \mathbf{m}_i}{\partial \mathbf{t}} = \frac{\partial \mathbf{m}_i}{\partial \mathbf{X}_i^c} = \left[\begin{matrix}\frac{f_x}{z^c} & 0 & - \frac{f_x  x^c}{z^c}\\0 & \frac{f_y}{z^c} & - \frac{f_y y^c}{(z^c)^{2}}\end{matrix}\right]
            \end{math}
        }
    \end{aligned}
\end{equation}
The gradients for rotation quaternion $\mathbf{q}$ is as follow:
\begin{equation}\label{eq:grad_q}
    \resizebox{0.85\linewidth}{!}{
        \begin{math}
            \begin{aligned}
                 & \frac{\partial \mathbf{X}_i^c}{\partial q_r} = 2\left[\begin{array}{ccc}
                                                                                 0       & - q_k y & q_j z  \\
                                                                                 q_k x   & 0       & -q_i z \\
                                                                                 - q_j x & q_i y   & 0
                                                                             \end{array}\right],
                 & \frac{\partial \mathbf{X}_i^c}{\partial q_i} = 2\left[\begin{array}{ccc}
                                                                                 0       & q_j y   & q_k z   \\
                                                                                 q_j x   & -2q_i y & - q_r z \\
                                                                                 - q_k x & q_r y   & -2q_i z
                                                                             \end{array}\right], \\
                 & \frac{\partial \mathbf{X}_i^c}{\partial q_j} = 2\left[\begin{array}{ccc}
                                                                                 - 2q_j x & q_i y & q_r z    \\
                                                                                 q_i x    & 0     & q_k z    \\
                                                                                 - q_r x  & q_k y & -2 q_j z
                                                                             \end{array}\right],
                 & \frac{\partial \mathbf{X}_i^c}{\partial q_k} = 2\left[\begin{array}{ccc}
                                                                                 - 2q_k x & - q_r y  & q_i z \\
                                                                                 q_r x    & - 2q_k y & q_j z \\
                                                                                 q_i x    & q_j y    & 0
                                                                             \end{array}\right]
            \end{aligned}
        \end{math}
     }
\end{equation}

\vspace{0.5ex}
\boldparagraph{Pose gradients back-propagation by $\frac{\partial\mathbf{\Sigma^\prime}}{\partial{\mathbf{P}}}$.} We first derive the gradients for $\mathbf{E} = \mathbf{JP}^{-1}= [e_{00}, e_{01}, e_{02}; e_{10}, e_{11}, e_{12}]$, where $\mathbf{J} = \frac{\partial \mathbf{m}_i}{\partial \mathbf{X}^c}$ is the Jacobian of $\mathbf{m}_i$ w.r.t. $\mathbf{X}^c$:

\begin{equation}\label{eq:grad_part_2}
    \resizebox{0.4\linewidth}{!}{
        \begin{math}
            \begin{aligned}
                \frac{\partial\mathbf{\Sigma^{\prime}}}{\partial{\mathbf{P}}} = \frac{\partial(\mathbf{JP}^{-1}\mathbf{\Sigma P}^{-T}\mathbf{J}^{T})}{\partial{\mathbf{P}}}
            \end{aligned}
        \end{math}
    }
\end{equation}
Then, we back-propagate the gradient to $\mathbf{E}$:
\begin{equation}\label{eq:grad_m}
    \resizebox{0.85\linewidth}{!}{
        \begin{math}
            \begin{aligned}
                 & \frac{\partial \text{vec}(\mathbf{\Sigma^{\prime}})}{\partial e_{00}} = \left[\begin{array}{c}
                                                                                                         2 e_{00} \Sigma_{00} + e_{01} \Sigma_{01} + e_{01} \Sigma_{10} + e_{02} \Sigma_{02} + e_{02} \Sigma_{20} \\
                                                                                                         e_{10} \Sigma_{00} + e_{11} \Sigma_{01} + e_{12} \Sigma_{02}                                             \\
                                                                                                         e_{10} \Sigma_{00} + e_{11} \Sigma_{10} + e_{12} \Sigma_{20}                                             \\
                                                                                                         0
                                                                                                     \end{array}\right], \\
                 & \frac{\partial \text{vec}(\mathbf{\Sigma^{\prime}})}{\partial e_{01}} = \left[\begin{array}{c}
                                                                                                         e_{00} \Sigma_{01} + e_{00} \Sigma_{10} + 2 e_{01} \Sigma_{11} + e_{02} \Sigma_{12} + e_{02} \Sigma_{21} \\
                                                                                                         e_{10} \Sigma_{10} + e_{11} \Sigma_{11} + e_{12} \Sigma_{12}                                             \\
                                                                                                         e_{10} \Sigma_{01} + e_{11} \Sigma_{11} + e_{12} \Sigma_{21}                                             \\
                                                                                                         0
                                                                                                     \end{array}\right], \\
                 & \frac{\partial \text{vec}(\mathbf{\Sigma^{\prime}})}{\partial e_{02}} = \left[\begin{array}{c}
                                                                                                         e_{00} \Sigma_{02} + e_{00} \Sigma_{20} + e_{01} \Sigma_{12} + e_{01} \Sigma_{21} + 2 e_{02} \Sigma_{22} \\
                                                                                                         e_{10} \Sigma_{20} + e_{11} \Sigma_{21} + e_{12} \Sigma_{22}                                             \\
                                                                                                         e_{10} \Sigma_{02} + e_{11} \Sigma_{12} + e_{12} \Sigma_{22}                                             \\
                                                                                                         0
                                                                                                     \end{array}\right], \\
                 & \frac{\partial \text{vec}(\mathbf{\Sigma^{\prime}})}{\partial e_{10}} = \left[\begin{array}{c}
                                                                                                         0                                                            \\
                                                                                                         e_{00} \Sigma_{00} + e_{01} \Sigma_{10} + e_{02} \Sigma_{20} \\
                                                                                                         e_{00} \Sigma_{00} + e_{01} \Sigma_{01} + e_{02} \Sigma_{02} \\
                                                                                                         2 e_{10} \Sigma_{00} + e_{11} \Sigma_{01} + e_{11} \Sigma_{10} + e_{12} \Sigma_{02} + e_{12} \Sigma_{20}
                                                                                                     \end{array}\right],  \\
                 & \frac{\partial \text{vec}(\mathbf{\Sigma^{\prime}})}{\partial e_{11}} = \left[\begin{array}{c}
                                                                                                         0                                                            \\
                                                                                                         e_{00} \Sigma_{01} + e_{01} \Sigma_{11} + e_{02} \Sigma_{21} \\
                                                                                                         e_{00} \Sigma_{10} + e_{01} \Sigma_{11} + e_{02} \Sigma_{12} \\
                                                                                                         e_{10} \Sigma_{01} + e_{10} \Sigma_{10} + 2 e_{11} \Sigma_{11} + e_{12} \Sigma_{12} + e_{12} \Sigma_{21}
                                                                                                     \end{array}\right],  \\
                 & \frac{\partial \text{vec}(\mathbf{\Sigma^{\prime}})}{\partial e_{12}} = \left[\begin{array}{c}
                                                                                                         0                                                            \\
                                                                                                         e_{00} \Sigma_{02} + e_{01} \Sigma_{12} + e_{02} \Sigma_{22} \\
                                                                                                         e_{00} \Sigma_{20} + e_{01} \Sigma_{21} + e_{02} \Sigma_{22} \\
                                                                                                         e_{10} \Sigma_{02} + e_{10} \Sigma_{20} + e_{11} \Sigma_{12} + e_{11} \Sigma_{21} + 2 e_{12} \Sigma_{22}
                                                                                                     \end{array}\right]
            \end{aligned}
        \end{math}
    }
\end{equation}

In the mapping process, oversized 3D Gaussians are controlled through the delete, split, and clone process, so the magnitude of the covariance $\mathbf{\Sigma}$ is small enough. When back-propagate the gradients to pose $\mathbf{P}$, the intermediate term in ~\cref{eq:grad_m} can be ignored.

\vspace{0.5ex}
\boldparagraph{Pose gradients back-propagation by depth supervision.}
Point-based depth alpha blending and color alpha blending share similarities; therefore, we implement the back-propagation of gradients for depth in the same manner for color:
\begin{equation}\label{eq:grad_depth}
    \resizebox{0.85\linewidth}{!}{
        \begin{math}
            \begin{aligned}
                 & \frac{\partial D}{\partial \alpha_i} = \begin{array}{c}
                                                              d_i \prod_{j=1}^{i-1}(1-\alpha_j) - \sum_{k=i+1}^{n} d_k \alpha_k \prod_{j=1, j \neq i}^{k-1} (1-\alpha_j)
                                                          \end{array}, \\
                 & \frac{\partial D}{\partial d_i}= \begin{array}{c}
                                                        \alpha_i  \prod_{j=1}^{i-1} \left(1-\alpha_j\right)
                                                    \end{array},
            \end{aligned}
        \end{math}
    }
\end{equation}
where $n$ is the number of the 3D Gaussian splats that affect the pixel in the rasterization.


\begin{figure*}[t]
    \begin{center}
        \includegraphics[width=0.85\linewidth]{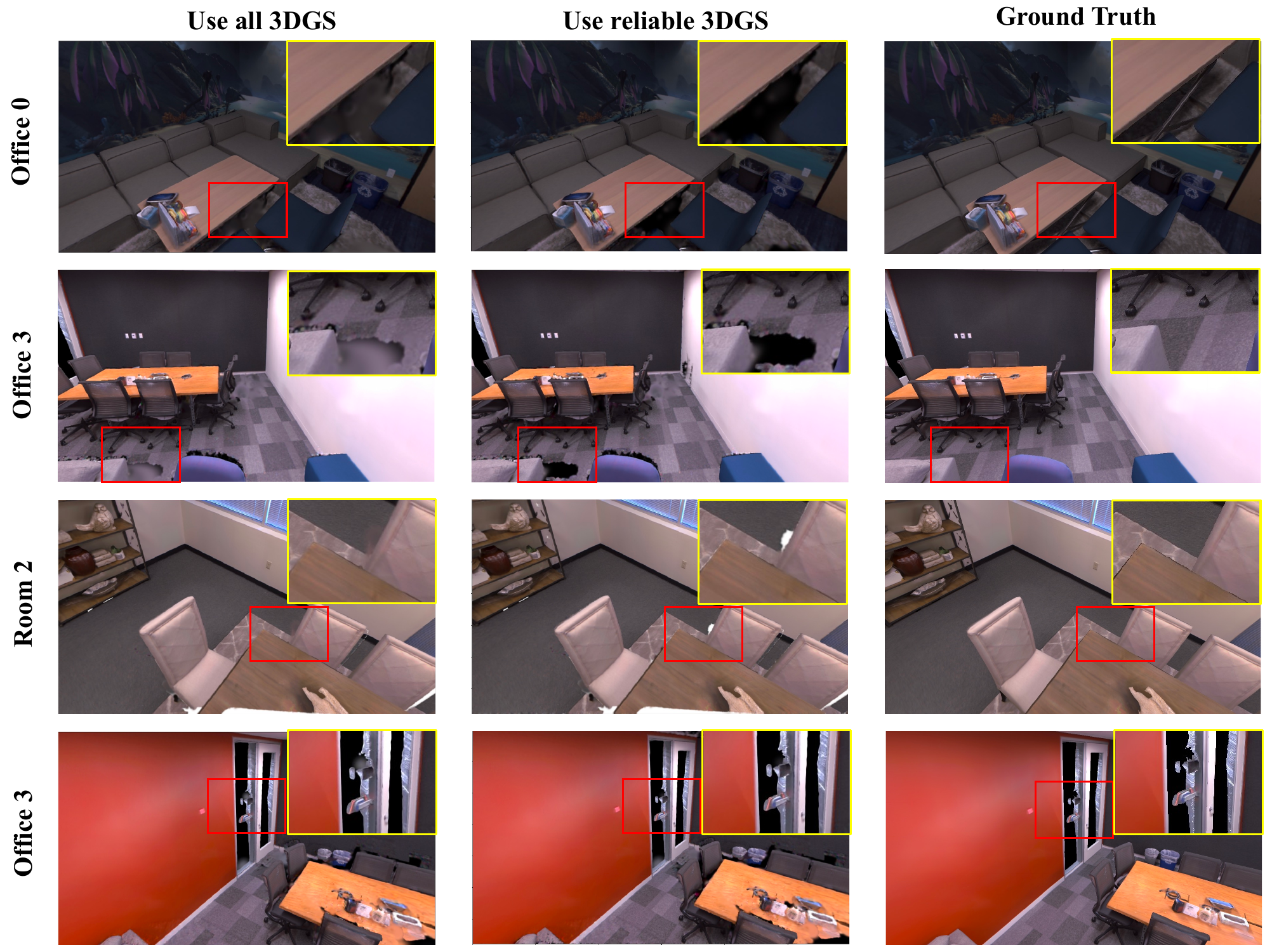}
    \end{center}
    \vspace{-2ex}
    \caption{Visualization of the rasterized result before and after selecting reliable Gaussians. \ours coarse-to-fine strategy can effectively remove the unreliable 3D Gaussians to obtain more precise tracking results. In the enlarged figure, it can be seen that the abnormal optimized 3D Gaussians are removed. Note that the pixel without any 3D Gaussian will not affect the gradients of the pose. The background color in \texttt{\#office 0} and  \texttt{\#office 3} is set to black, while in \texttt{\#room 0} is white.}
    \vspace{-2ex}
    \label{fig:c2f_vis}
\end{figure*}

\section{Coarse-to-fine Pose Optimization Algorithm}
\label{sec:c2f_pose}
Our coarse-to-fine pose optimization algorithms are summarized in~\cref{alg:optimization}.

Our algorithm is designed for splatting-based pose estimation, which uses $\alpha$-blending on 3D Gaussians in strict front-to-back order rather than the volume rendering technique used by current NeRF-Based SLAM~\cite{Zhu2021NICESLAMNI, Wang2023CoSLAMJC, Johari2022ESLAMED, pointslam, zhang2023goslam, Yang2022VoxFusionDT, Sucar2021iMAPIM}.

 Visualization of the unreliable area removed in the fine stage is shown in~\cref{fig:c2f_vis}.
In the $\texttt{\#Use all 3DGS}$ column, we can see artifacts within the red bounding boxes resulting from improperly optimized positions in the past mapping process. These artifacts significantly impact the accuracy of pose estimation, as they introduce excessive optimization errors in the loss function.
Our proposed reliable 3D Gaussians selection in coarse-to-fine strategy, as shown in the second column, filters the unreliable 3D Gaussians. The remaining background(black in $\texttt{\#Office 0}$, $\texttt{\#Office 3}$ and white in $\texttt{\#Room 0}$) on the image plane would be ignored while back-propagating gradients to pose. Details about tracking and mapping settings can be seen in~\cref{sec:add_implement_detail}.

\begin{algorithm}[h]
    \caption{\textbf{Coarse-to-fine pose optimization}\\
        $W$, $H$: width and height of the input images}
    \label{alg:optimization}
    \begin{algorithmic}
        \State $\mathbf{P}^{(0)}$ $\gets$ $\mathbf{P}_{t-1}$, $\mathbf{P}_{t-2}$ \Comment{Init Pose}
        \State $i_c, i_f \gets 0$	\Comment{Coarse and Fine Iteration Count}
        \For{$i_c < T_{c}$}
        \State $\hat{I}_c \gets$ Rasterize($\mathbf{G}$, $\mathbf{P}^{(i_c)}$, $\frac{1}{2}H$, $\frac{1}{2}W$)	\\
        \Comment{Coarse Render}
        \State $L \gets Loss(I_c, \hat{I}_c) $ \Comment{Loss}
        \State $\mathbf{P}^{(i_{c}+1)}$ $\gets$ Adam($\nabla L$) \Comment{Backprop \& Step}
        \State $i_c \gets i_c+1$
        \EndFor

        \State $\mathbf{G}_{selected}$ $\gets$ ReliableGS($\mathbf{G}, \mathbf{P}^{(i_{c}+1)}, D, \varepsilon  $)\\ \Comment{Select Reliable $\mathbf{G}$}
        \For{$i_f < T_{f}$}
        \State $\hat{I}_f \gets$ Rasterize($\mathbf{G}_{selected}$, $\mathbf{P}^{(i_c+i_f)}$, $H$, $W$)	\\
        \Comment{Fine Render}
        \State $L \gets Loss(I_f, \hat{I}_f) $ \Comment{Loss}
        \State $\mathbf{P}^{(i_{c}+{i_{f}}+1)}$ $\gets$ Adam($\nabla L$) \Comment{Backprop \& Step}
        \State $i_f \gets i_f+1$
        \EndFor
    \end{algorithmic}
\end{algorithm}

\begin{table*}[t]
    \centering
    \caption{Ablation study of tracking, mapping, and rendering performance on whole Replica dataset~\cite{straub2019replica}. We present detailed ablation results on the entire Replica dataset to demonstrate the significant advantages of our proposed modules.   Our method introduces a novel approach that combines coarse-to-fine pose estimation with an adaptive 3D Gaussian expansion strategy.   This comprehensive methodology successfully increases the render, tracking, and mapping quality.}
    \footnotesize
    \setlength{\tabcolsep}{2.65pt}
    \renewcommand{\arraystretch}{1}
    \begin{tabularx}{\linewidth}{l|lccccccccc}
        \hline
        Method & Metric                     & \texttt{Room 0} & \texttt{Room 1} & \texttt{Room 2} & \texttt{Office 0} & \texttt{Office 1} & \texttt{Office 2} & \texttt{Office 3} & \texttt{Office 4} & \texttt{Avg.} \\
        \hline
        \multirow{8}{*}{w/o delete in mapping}
               & ATE [cm] $\downarrow$      & \rd 0.58        & \nd 0.63        & \rd 0.57        & 0.76              & \fs 0.38          & 0.61              & 0.62              & 0.87              & \nd 0.63      \\
               & Depth L1 [cm] $\downarrow$ & 1.68            & 1.17            & \rd 1.92        & 1.30              & 1.46              & 1.82              & 1.79              & 1.83              & \rd 1.62      \\
               & Precision [\%]$\uparrow$   & 53.55           & 68.76           & \rd 50.58       & 67.03             & \rd 77.52         & 59.45             & 53.37             & 49.03             & 59.91         \\
               & Recall [\%]$\uparrow$      & 49.32           & 61.88           & \rd 45.57       & 61.60             & 65.67             & 50.15             & 46.09             & 42.80             & 52.89         \\
               & F1 [\%]$\uparrow$          & 51.35           & 65.14           & \rd 47.94       & 64.20             & \rd 71.10         & 54.41             & 49.46             & 45.70             & 56.16         \\
               & PSNR [dB] $\uparrow$       & \nd 31.22       & \fs 33.33       & \fs 33.58       & 38.17             & 39.97             & 30.77             & \fs 32.04         & 34.86             & \nd 34.24     \\
               & SSIM $\uparrow$            & \nd 0.967       & \fs 0.975       & \fs 0.977       & \rd 0.984         & \rd 0.990         & \nd 0.974         & \nd 0.969         & \nd 0.980         & \nd 0.977     \\
               & LPIPS $\downarrow$         & \fs 0.094       & \fs 0.075       & \fs 0.086       & 0.053             & \rd 0.046         & \rd 0.096         & \fs 0.100         & \fs 0.080         & \nd 0.079     \\
        \hline
        \multirow{8}{*}{Coarse in tracking}
               & ATE [cm] $\downarrow$      & 0.91            & 0.87            & \nd 0.52        & \rd 0.71          & 0.65              & \nd 0.56          & \rd 0.50          & \nd 0.71          & \rd 0.68      \\
               & Depth L1 [cm] $\downarrow$ & \rd 1.48        & \rd 0.94        & \nd 1.47        & \nd 0.84          & \nd 0.97          & \rd 1.52          & \rd 1.58          & \rd 1.28          & \nd 1.26      \\
               & Precision [\%]$\uparrow$   & \rd 59.68       & \rd 70.51       & \nd 62.66       & \nd 83.11         & \fs 87.79         & \rd 66.85         & \rd 61.34         & \rd 66.55         & \nd 69.81     \\
               & Recall [\%]$\uparrow$      & \rd 57.54       & \rd 64.98       & \nd 57.58       & \nd 76.36         & \nd 74.58         & \rd 58.25         & \nd 54.07         & \rd 59.01         & \nd 62.80     \\
               & F1 \%]$\uparrow$           & \rd 56.50       & \rd 67.63       & \nd 60.01       & \nd 79.59         & \nd 80.22         & \rd 62.25         & \rd 57.48         & \rd 62.55         & \nd 65.78     \\
               & PSNR [dB] $\uparrow$       & 29.13           & 32.08           & \nd 33.12       & \nd 38.62         & \nd 40.69         & \rd 32.02         & \nd 32.02         & \nd 35.05         & \rd 34.09     \\
               & SSIM $\uparrow$            & 0.954           & \nd 0.970       & \nd 0.971       & \fs 0.986         & \fs 0.993         & \fs 0.978         & \rd 0.967         & \nd 0.980         & \rd 0.975     \\
               & LPIPS $\downarrow$         & \nd 0.120       & \rd 0.085       & \nd 0.093       & \nd 0.051         & \nd 0.037         & 0.097             & \rd 0.117         & \rd 0.089         & \rd 0.086     \\
        \hline
        \multirow{8}{*}{Fine in tracking}
               & ATE [cm] $\downarrow$      & \nd 0.49        & \rd 0.82        & 5.59            & \nd 0.69          & \rd 0.57          & \fs 0.55          & \fs 0.40          & \rd 0.74          & 1.23          \\
               & Depth L1 [cm] $\downarrow$ & \nd 1.39        & \nd 0.91        & 5.84            & \rd 0.87          & \rd 1.26          & \nd 1.49          & \nd 1.55          & \nd 1.27          & 1.82          \\
               & Precision [\%]$\uparrow$   & \nd 62.61       & \nd 73.71       & 16.45           & \rd 81.63         & \rd 72.86         & \nd 69.08         & \nd 61.88         & \nd 67.05         & \rd 63.16     \\
               & Recall [\%] $\uparrow$     & \nd 59.18       & \nd 67.68       & 15.53           & \rd 75.53         & 64.77             & \nd 59.73         & \nd 54.40         & \nd 59.45         & \rd 57.03     \\
               & F1 [\%] $\uparrow$         & \nd 61.29       & \nd 70.57       & 15.98           & \rd 78.46         & 68.57             & \nd 64.06         & \nd 57.90         & \nd 63.02         & \rd 59.98     \\
               & PSNR [dB] $\uparrow$       & \rd 30.84       & \nd 33.247      & 27.25           & \rd 38.41         & \rd 40.46         & \nd 32.13         & \rd 32.03         & \nd 35.18         & 33.69         \\
               & SSIM $\uparrow$            & \rd 0.964       & \fs 0.975       & \rd 0.901       & \nd 0.985         & \nd 0.992         & \fs 0.978         & 0.966             & \nd 0.980         & 0.968         \\
               & LPIPS $\downarrow$         & \rd 0.096       & \nd 0.076       & \rd 0.188       & \rd 0.052         & \nd 0.037         & \nd 0.095         & 0.119             & \rd 0.089         & 0.094         \\
        \hline
        \multirow{8}{*}{Our full}
               & ATE [cm] $\downarrow$      & \fs 0.48        & \fs 0.53        & \fs 0.33        & \fs 0.52          & \nd 0.41          & \rd 0.59          & \nd 0.46          & \fs 0.70          & \fs 0.50      \\
               & Depth L1 [cm] $\downarrow$ & \fs 1.31        & \fs 0.82        & \fs 1.26        & \fs 0.81          & \fs 0.96          & \fs 1.41          & \fs 1.53          & \fs 1.08          & \fs 1.15      \\
               & Precision [\%]$\uparrow$   & \fs 64.58       & \fs 83.11       & \fs 70.13       & \fs 83.43         & \nd 87.77         & \fs 70.91         & \fs 63.18         & \fs 68.88         & \fs 74.09     \\
               & Recall [\%]$\uparrow$      & \fs 61.29       & \fs 76.83       & \fs 63.84       & \fs 76.90         & \fs 76.15         & \fs 61.63         & \fs 62.91         & \fs 61.50         & \fs 67.63     \\
               & F1 [\%]$\uparrow$          & \fs 62.89       & \fs 79.85       & \fs 66.84       & \fs 80.03         & \fs 81.55         & \fs 65.95         & \fs 59.17         & \fs 64.98         & \fs 70.16     \\
               & PSNR [dB] $\uparrow$       & \fs 31.56       & \rd 32.86       & \rd 32.59       & \fs 38.70         & \fs 41.17         & \fs 32.36         & \rd 32.03         & \fs 35.23         & \fs 34.56     \\
               & SSIM $\uparrow$            & \fs 0.968       & \nd 0.973       & \nd 0.971       & \fs 0.986         & \fs 0.993         & \fs 0.978         & \fs 0.970         & \fs 0.981         & \fs 0.978     \\
               & LPIPS $\downarrow$         & \fs 0.094       & \fs 0.075       & \nd 0.093       & \fs 0.050         & \fs 0.033         & \fs 0.009         & \nd 0.110         & \nd 0.088         & \fs 0.069     \\
        \hline
    \end{tabularx}
    \vspace{-2ex}
    \label{tab:abl_replica}
\end{table*}

\section{Additional Performance Comparison}
\label{sec:performance}
\boldparagraph{Render performance on TUM-RGBD dataset.}
As shown in~\cref{fig:add_viz_tum}, we compare our method to current state-of-the-art NeRF-Based SLAM method CoSLAM~\cite{Wang2023CoSLAMJC}, ESLAM~\cite{Johari2022ESLAMED}, Point-SLAM~\cite{pointslam} and ground truth image. We showcase the render quality of different methods using the final reconstructed environment model, presented in descending order from top to bottom.
Our GS-SLAM provides the clearest results, particularly evident in the complex $\texttt{\#fr3\_office}$ scene, displaying a higher richness in detail information, indicating its superiority in handling details and edges.

\boldparagraph{Render performance on Replica dataset.}
As shown in~\cref{fig:add_viz_replica}, our GS-SLAM method performs superior in all tested scenarios.  It delivers renderings with clear, well-defined edges and richly textured surfaces that closely align with the ground truth.  This suggests that GS-SLAM not only captures the geometric detail with high precision but also accurately reconstructs the textural information of the scenes.  The method's ability to render sharp images even in regions with complex textures and lighting conditions underscores its potential for accurate 3D environment mapping.

\section{Additional Ablation Results}
\label{sec:full_ab}
\noindent{\textbf{More detailed ablation experiments.}}
In \cref{tab:abl_replica}, we present our ablation study that demonstrates the effectiveness of our proposed approach on the comprehensive Replica dataset. The experiment contrasts different module arrangements, including disabling our Gaussian delete method in mapping, using only coarse images in tracking, using fine images in tracking, and using our fully integrated method. The results distinctly highlight the superiority of our complete methodology, particularly evident in improved metrics across the board, from ATE to LPIPS. This confirms the benefits of our coarse-to-fine pose estimation and adaptive 3D Gaussian expansion strategy.

\begin{table}[t]
    \vspace{0ex}
    \centering
    \scriptsize
    \caption{Ablation of the depth supervision on Replica \texttt{\#Room0}.}
    \resizebox{0.48\textwidth}{!}{
        \setlength{\tabcolsep}{2.5pt}
        \begin{tabular}{lcccccccc}
            \toprule
            \multirow{2}{*}{Setting} & \multicolumn{8}{c}{\texttt{\#Room0}}                                                                                                                                                                                                                                                                                                                  \\\cline{2-9}
                                     & \multirow{2}{*}[0.5ex]{ATE$\downarrow$} & \multirow{2}{*}[0.5ex]{Depth L1$\downarrow$} & \multirow{2}{*}[0.5ex]{Precision$\uparrow$} & \multirow{2}{*}[0.5ex]{Recall $\uparrow$} & \multirow{2}{*}[0.5ex]{F1$\uparrow$} & \multirow{2}{*}[0.5ex]{PSNR$\uparrow$} & \multirow{2}{*}[0.5ex]{SSIM$\uparrow$} & \multirow{2}{*}[0.5ex]{LPIPS$\downarrow$} \\
            \midrule
            w/o Depth                & 0.80                                    & 3.21                                         & 14.28                                       & 15.01                                     & 14.63                                & 29.76                                  & 0.956                                  & 0.107                                     \\
            \midrule
            w/ Depth                 & \fs{0.48}                           & \fs{1.31}                                & \fs{64.58}                              & \fs{61.29}                            & \fs{62.89}                       & \fs{31.56}                         & \fs{0.968}                         & \fs{0.094}                            \\
            \bottomrule
        \end{tabular}
    }
    \label{tab:ab_depth}
    \vspace{-5ex}
\end{table}

\begin{figure*}[t]
    \vspace{-4ex}
    \centering
    {\footnotesize
        \setlength{\tabcolsep}{0.9pt}
        \renewcommand{\arraystretch}{0}
        \newcommand{\sz}{0.07}
        \begin{tabular}{ccccc}
                                                                     & \texttt{\#Room 2}                                                                                                        & \texttt{\#Office 0}                                                                                                        & \texttt{\#Office 3}                                                                                                        & \texttt{\#Office 4}                                                                                                        \\

            \rotatebox[origin=c]{90}{CoSLAM~\cite{Wang2023CoSLAMJC}} & \includegraphics[valign=c,height=\sz\linewidth]{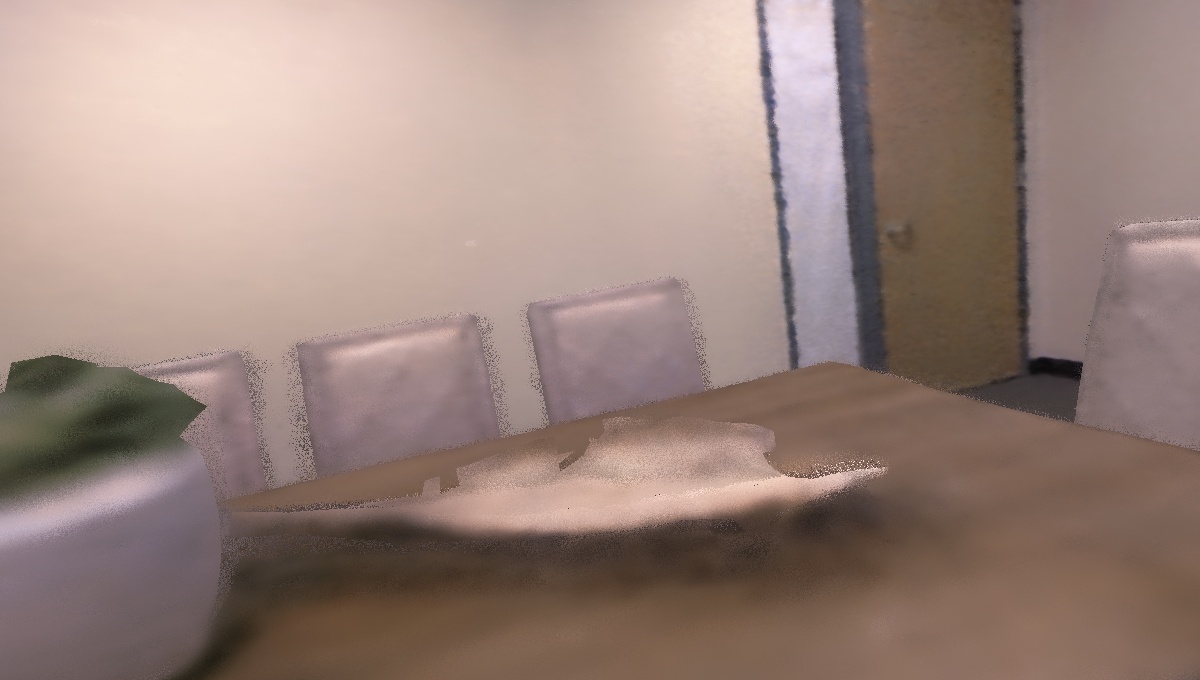}    & \includegraphics[valign=c,height=\sz\linewidth]{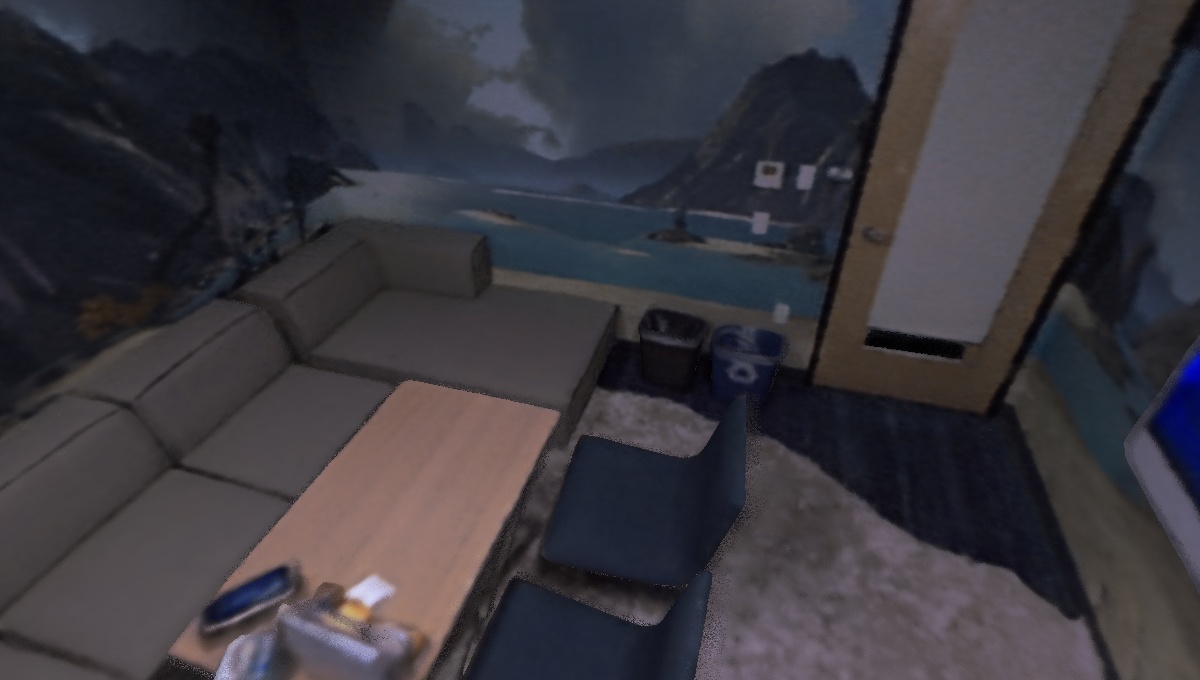}    & \includegraphics[valign=c,height=\sz\linewidth]{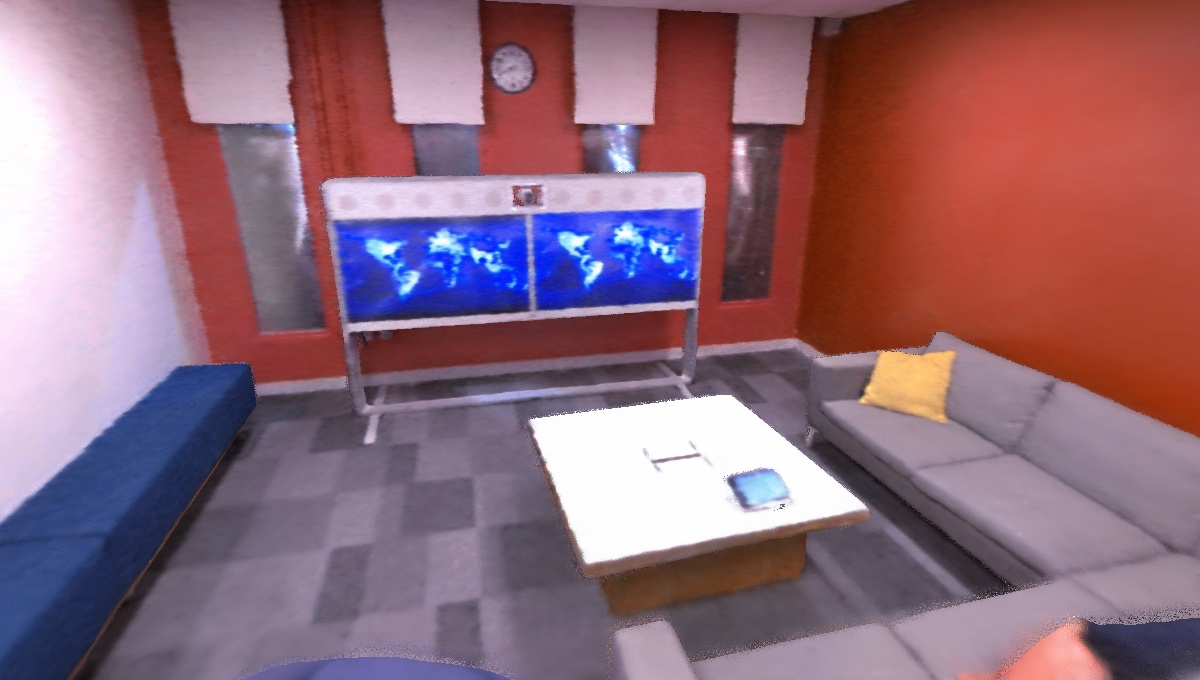}    & \includegraphics[valign=c,height=\sz\linewidth]{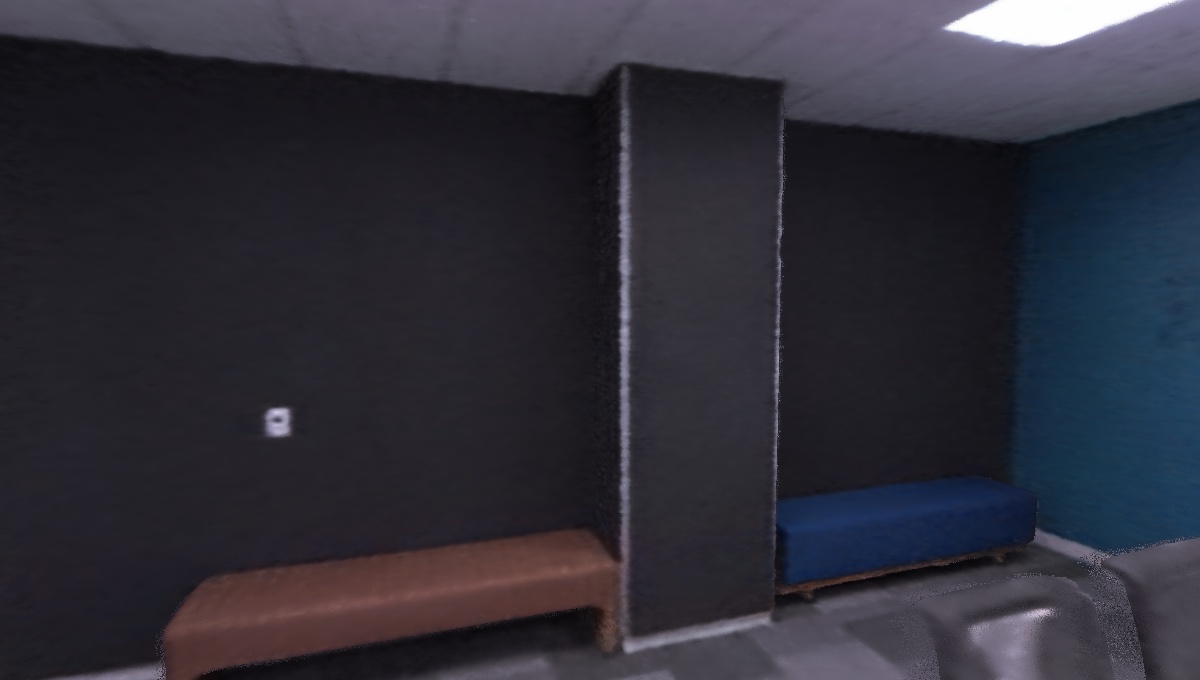}    \\

            \rotatebox[origin=c]{90}{ESLAM~\cite{Johari2022ESLAMED}} & \includegraphics[valign=c,height=\sz\linewidth]{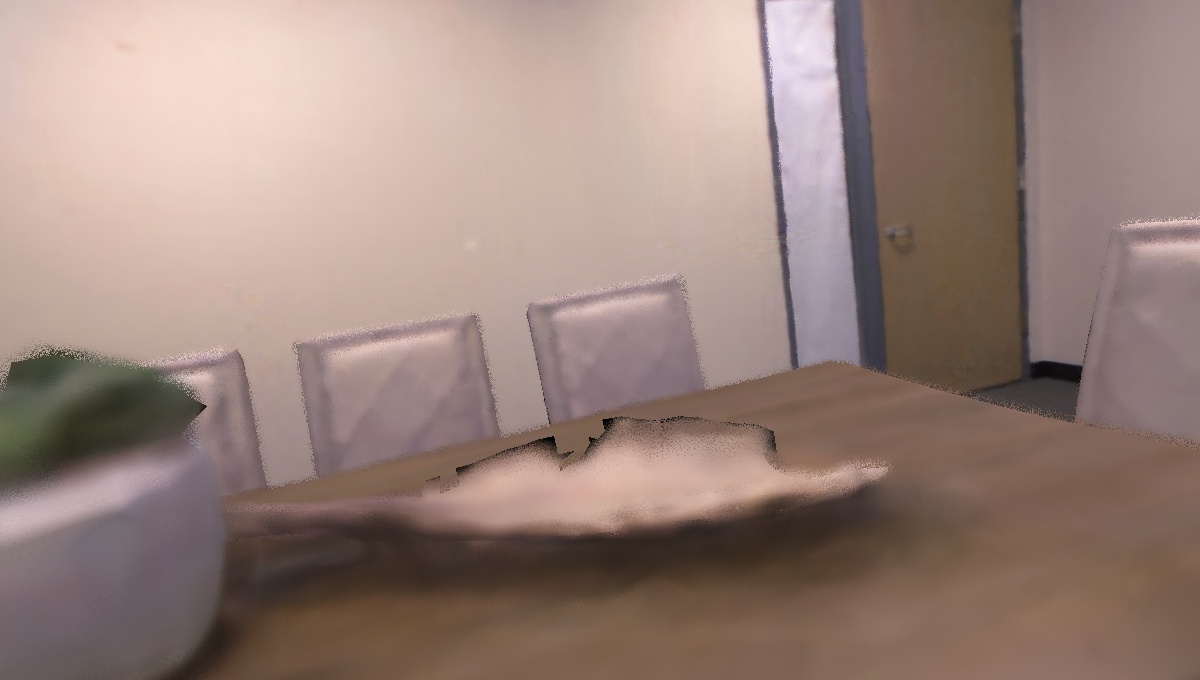}     & \includegraphics[valign=c,height=\sz\linewidth]{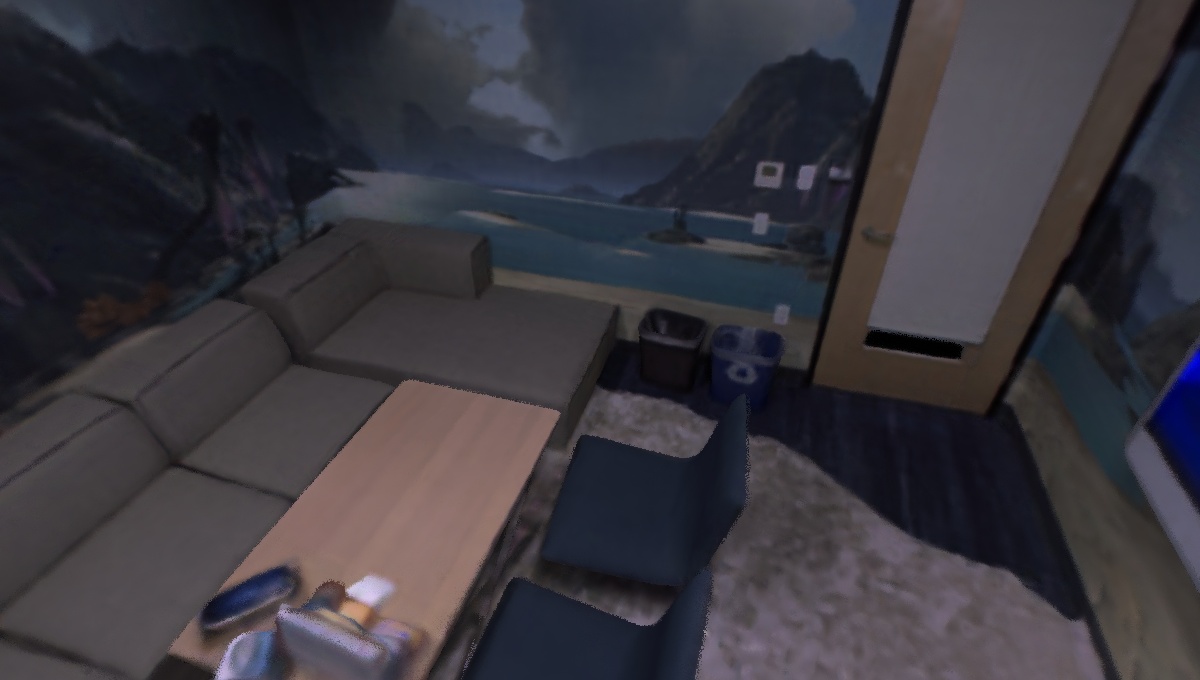}     & \includegraphics[valign=c,height=\sz\linewidth]{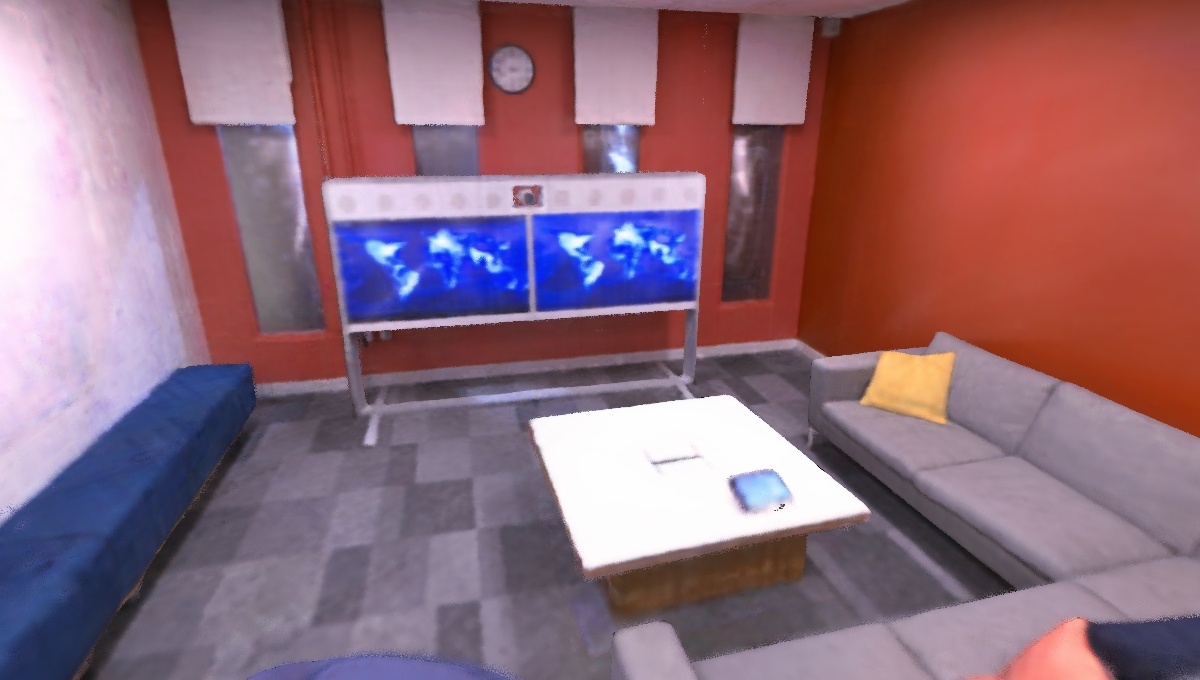}     & \includegraphics[valign=c,height=\sz\linewidth]{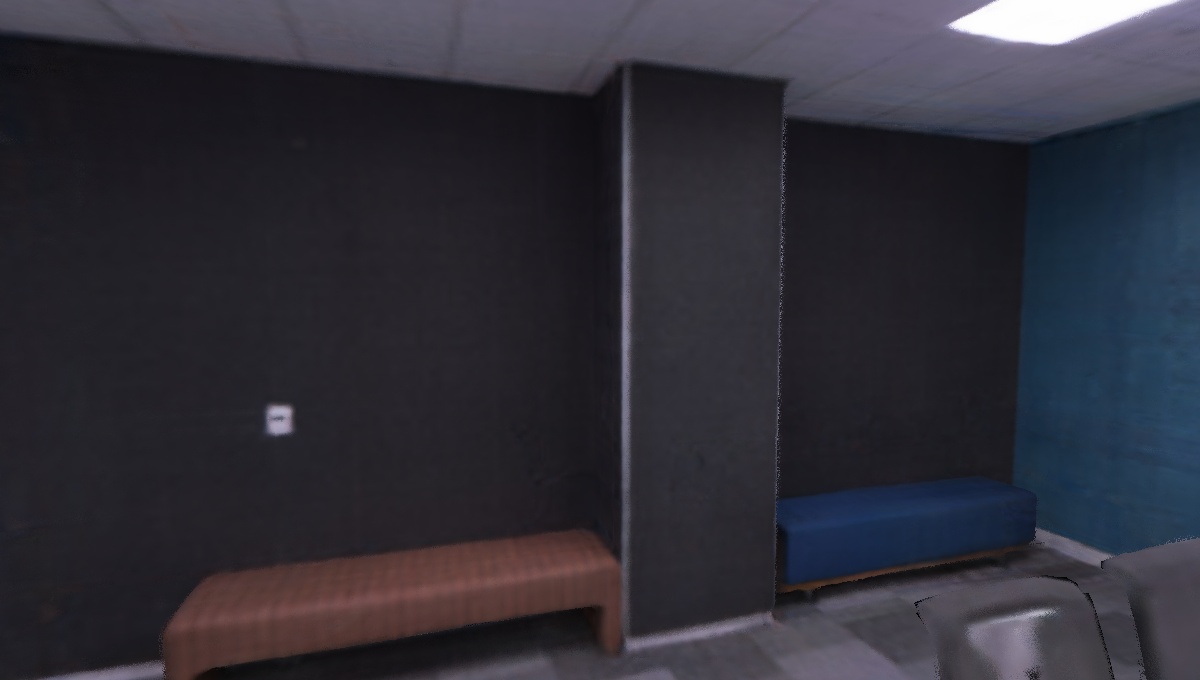}     \\

            \rotatebox[origin=c]{90}{Point-SLAM~\cite{pointslam}}    & \includegraphics[valign=c,height=\sz\linewidth]{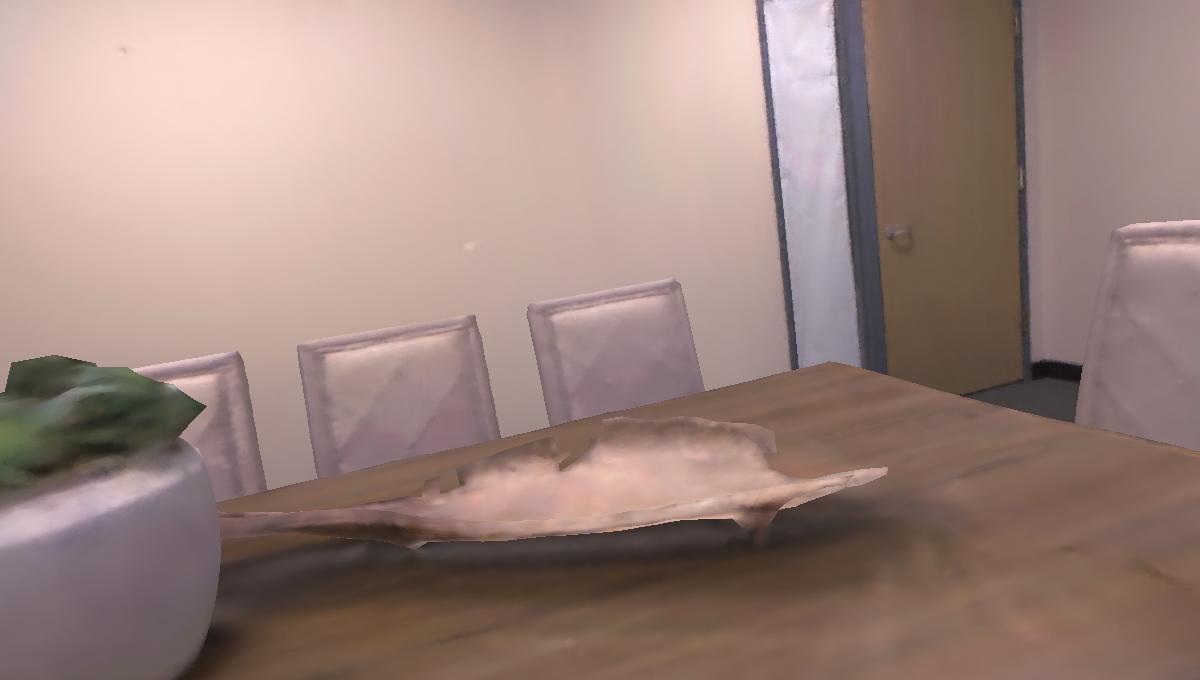} & \includegraphics[valign=c,height=\sz\linewidth]{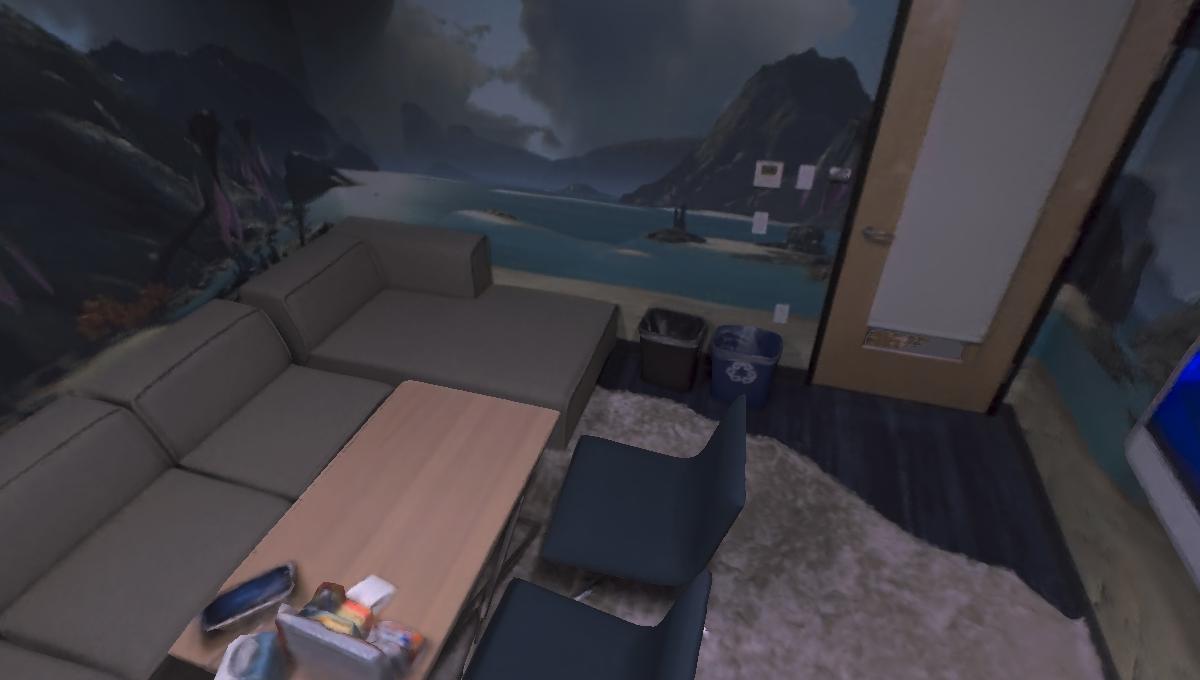} & \includegraphics[valign=c,height=\sz\linewidth]{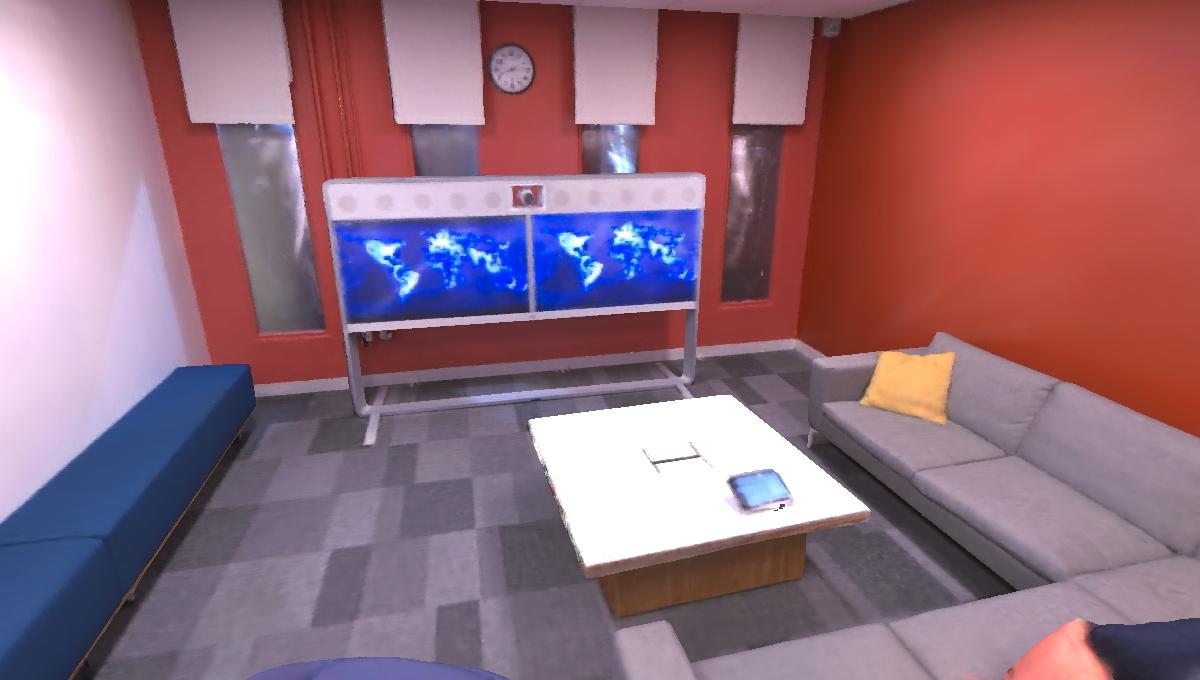} & \includegraphics[valign=c,height=\sz\linewidth]{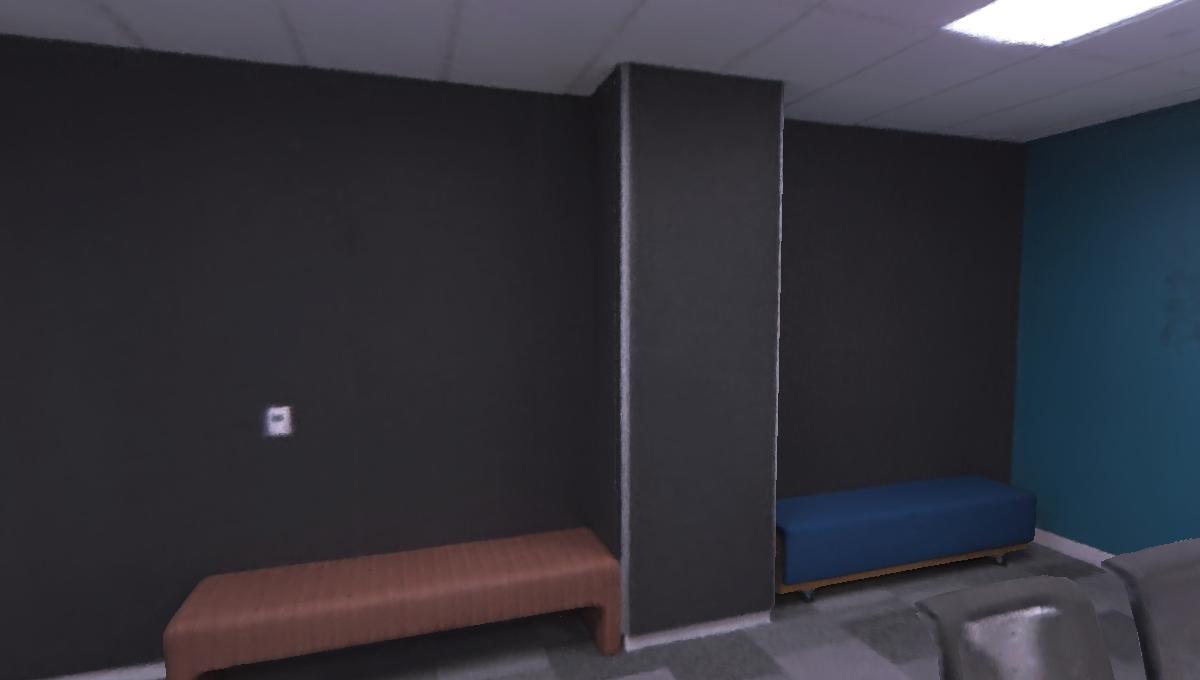} \\

            \rotatebox[origin=c]{90}{GS-SLAM (Ours)}                 & \includegraphics[valign=c,height=\sz\linewidth]{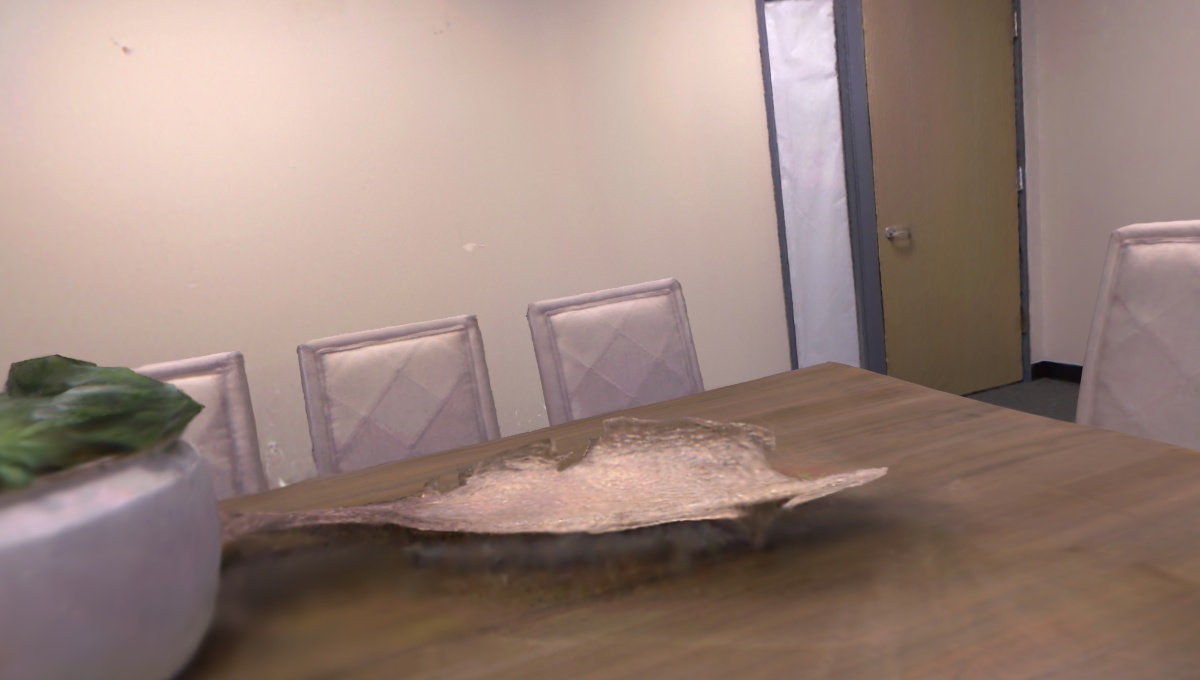}    & \includegraphics[valign=c,height=\sz\linewidth]{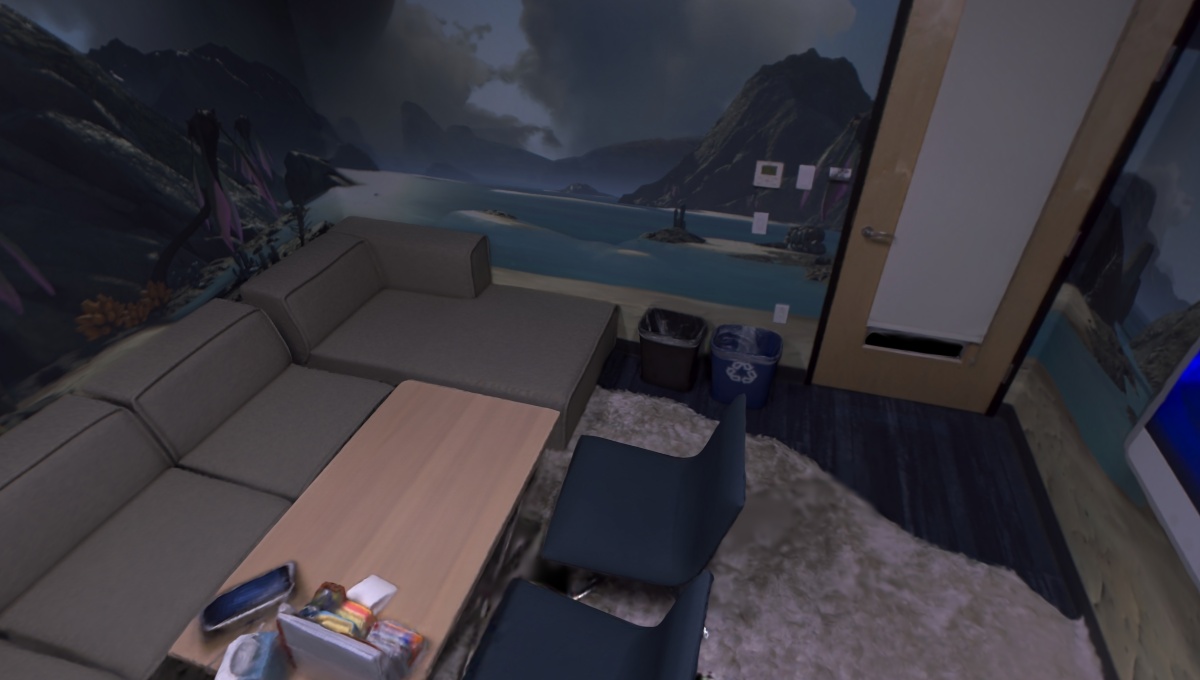}    & \includegraphics[valign=c,height=\sz\linewidth]{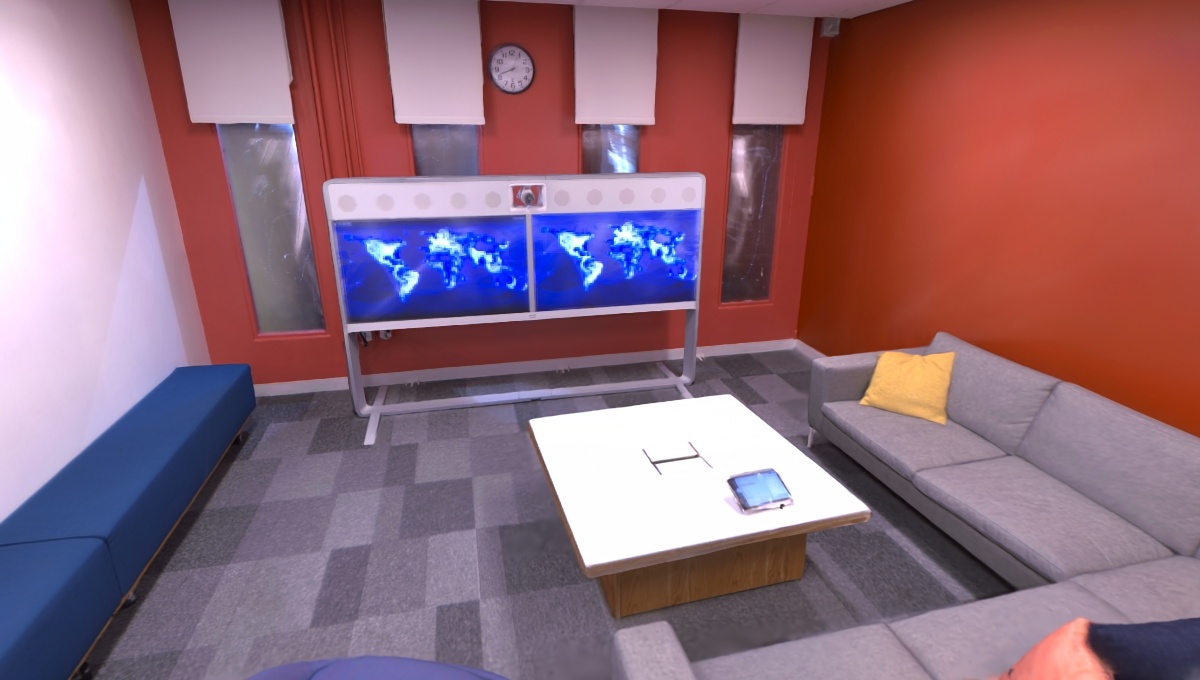}    & \includegraphics[valign=c,height=\sz\linewidth]{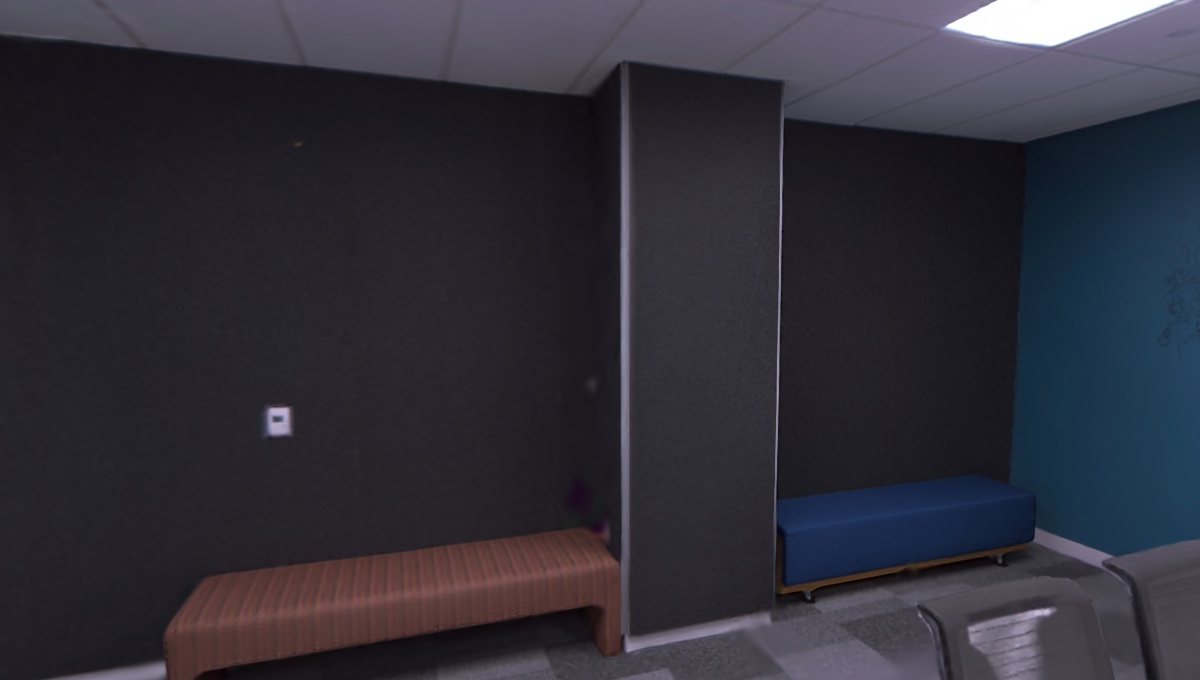}    \\

            \rotatebox[origin=c]{90}{Ground Truth}                   & \includegraphics[valign=c,height=\sz\linewidth]{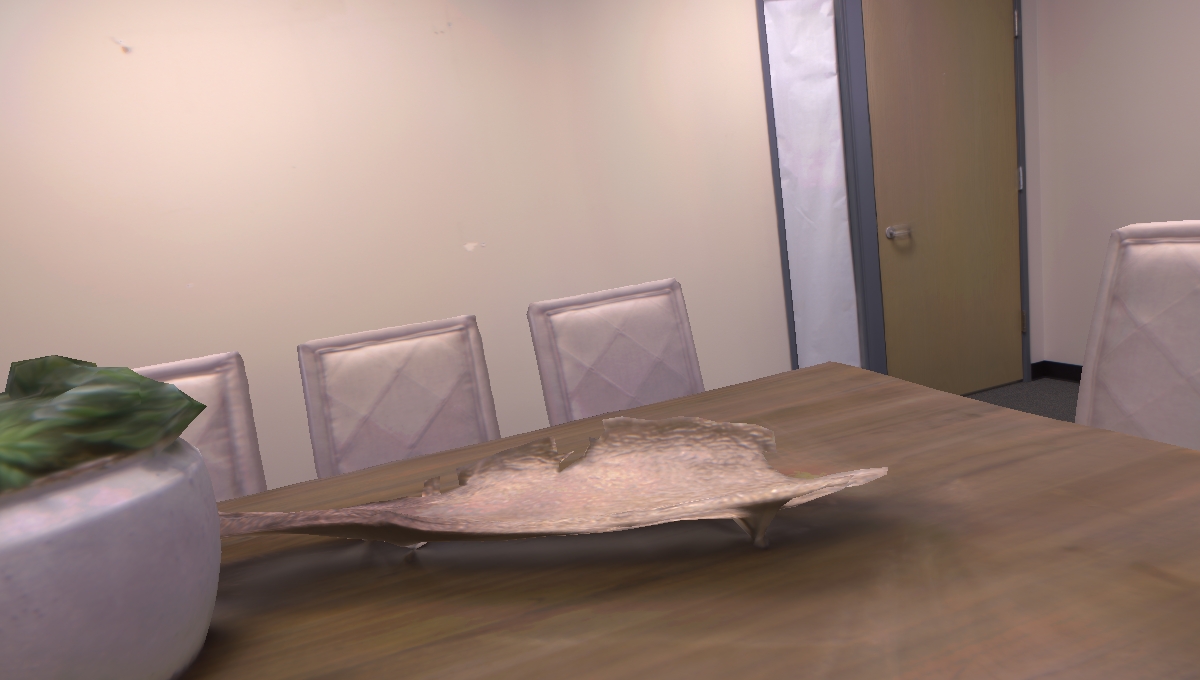}         & \includegraphics[valign=c,height=\sz\linewidth]{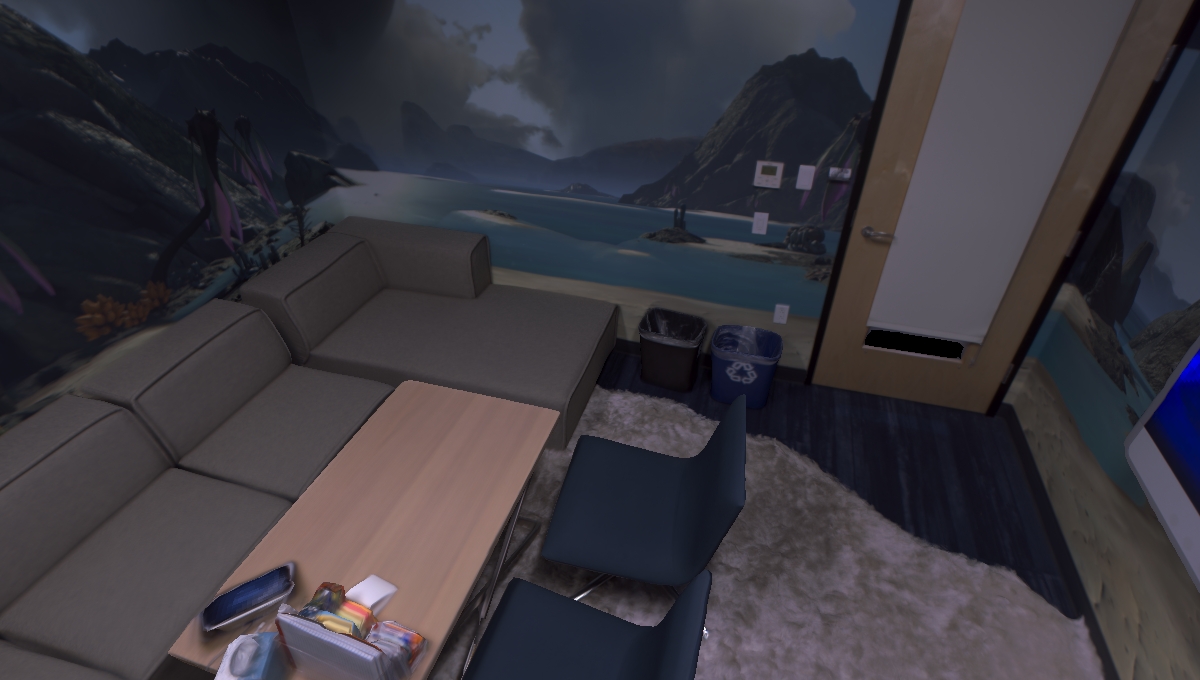}         & \includegraphics[valign=c,height=\sz\linewidth]{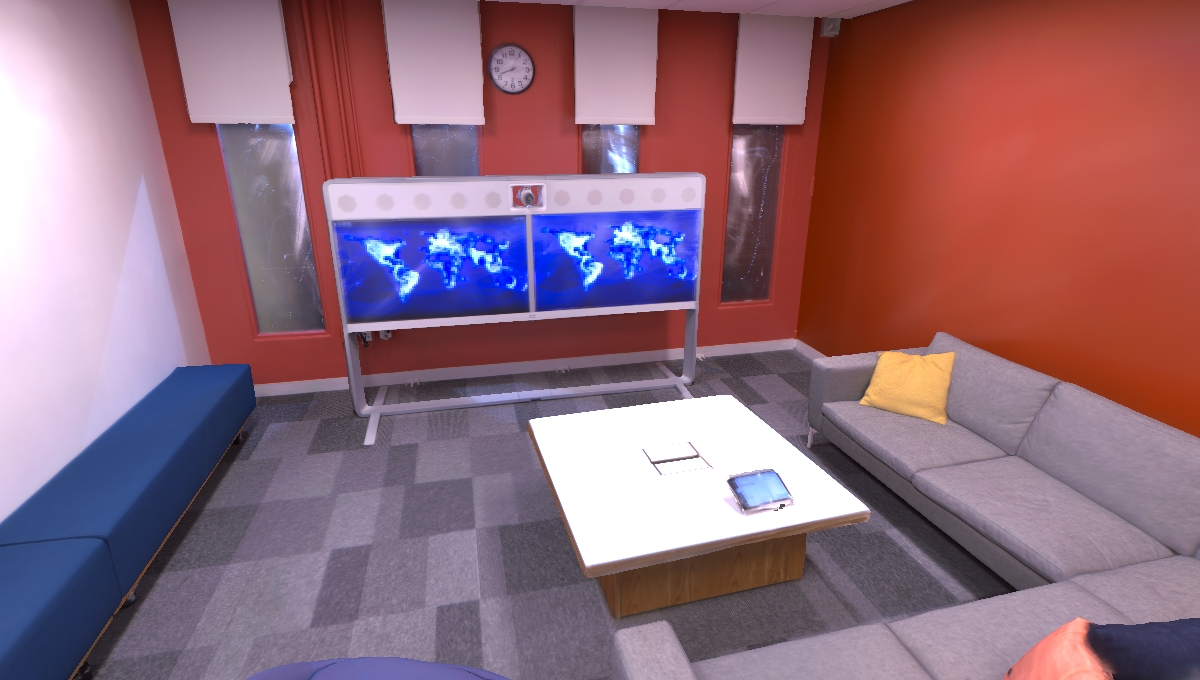}         & \includegraphics[valign=c,height=\sz\linewidth]{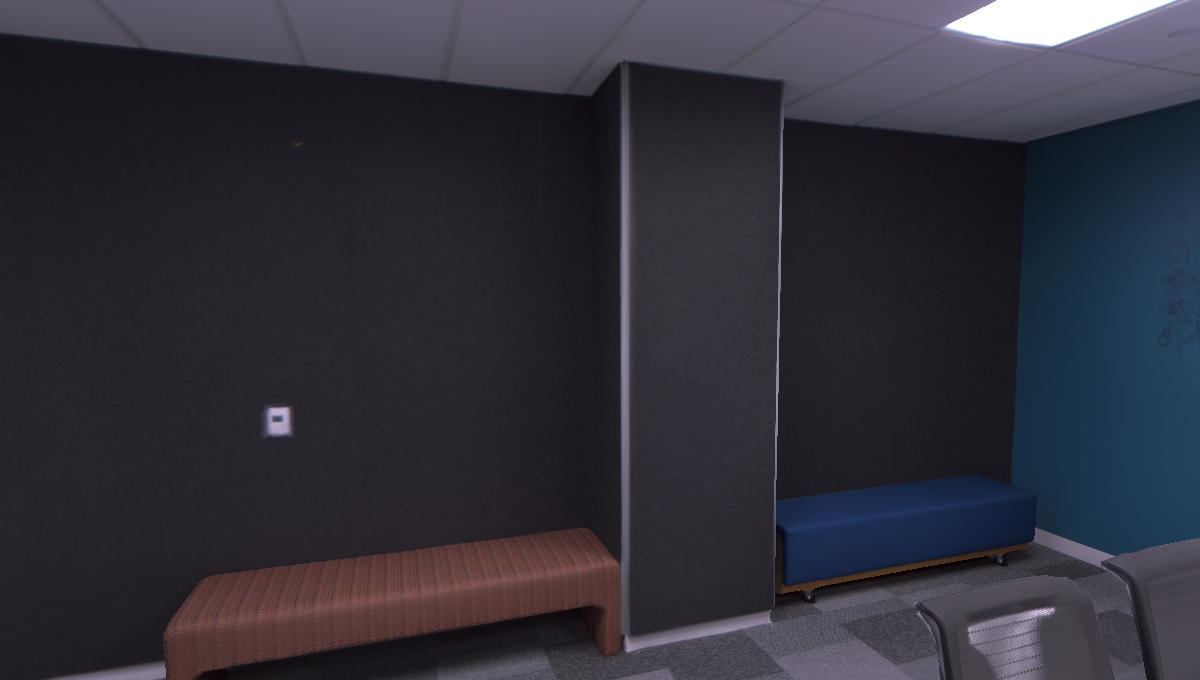}         \\

        \end{tabular}
    }
    \caption{Additional render result comparison on Replica~\cite{straub2019replica}. Our GS-SLAM can achieve more clear edges and better results in regions with rich texture than the previous SOTA methods.}
    \label{fig:add_viz_replica}
\end{figure*}

\noindent{\textbf{Effect of adaptive 3D Gaussian expansion strategy.}}
~\cref{fig:ba_delete} demonstrates the reconstructed mesh with and without our adaptive 3D Gaussian expansion strategy. The comparison is based on the Replica dataset $\texttt{Room0}$ subsequence. The left side displays a more coherent surface mesh due to our expansion strategy, while the right side lacks this delete strategy and results in less accuracy in reconstruction.

\noindent{\textbf{Effect of depth supervision.}}
\cref{tab:ab_depth} illustrates quantitative evaluation using depth supervision in mapping. In contrast to the original color-only supervision in~\cite{kerbl3Dgaussians}, the depth supervision can significantly improve the tracking and mapping performance by providing accurate geometry constraints for the optimization. Our implementation with depth achieves better tracking ATE of 0.48, mapping precision of 64.58, and rendering PSNR of 31.56 compared with the implementation without depth supervision.

\section{Additional Implementation Details}
\label{sec:add_implement_detail}
\boldparagraph{Mapping hyper-parameters.}
The 3D Gaussian representation and pose are trained using Adam optimizer with initialized position learning rate $lr_{\mathbf{X}_{init}}=1.6e^{-5}$, final position learning rate $lr_{\mathbf{X}_{final}}=1.7e^{-7}$, max attenuation steps $100$.
Other Spherical Harmonics coefficients learning rate $lr_{\boldsymbol{Y}}$, opacity learning rate $lr_\Lambda$, scaling learning rate $lr_{\mathbf{S}}$ and rotation learning rate $lr_{\mathbf{R}}$ set to $5e^{-4}$, $1e^{-2}$, $2e^{-4}$, $4e^{-5}$ respectively.
In all experiments, we set the photometric loss weighting $0.8$,  geometric loss weighting $0.3$, and keyframe window size $K=10$.  
In the mapping process, we densify the 3D Gaussians every 10 iterations before the first 70 iterations in a total of 100 iterations. And the scale and grad threshold for clone or split is set to $0.02m$ and $0.002$.
For the stability of the optimization, first-order coefficients of spherical harmonics coefficients are only optimized in bundle adjustment. Note that we only optimize the camera pose in the latter half of the iterations due to the adverse impact of improper 3D Gaussians on optimization. Despite this, there are still negative optimizations for camera poses at some point. 
In addition, in all TUM-RGBD sequences and Replica \texttt{Office} subsequence, we set the background color to black, while in other \texttt{Room} subsequence, we set the background to white. In the Replica dataset, we use 10 iterations for tracking and 100 iterations for mapping with max keyframe interval $\mu_{k}=30$, while in the challenging TUM RGB-D dataset, we use 30 iterations for tracking, with max keyframe interval $\mu_{k}= 5$.

\begin{figure}[htbp]
    \vspace{-2.5ex}
    \begin{center}
        \includegraphics[width=1\linewidth]{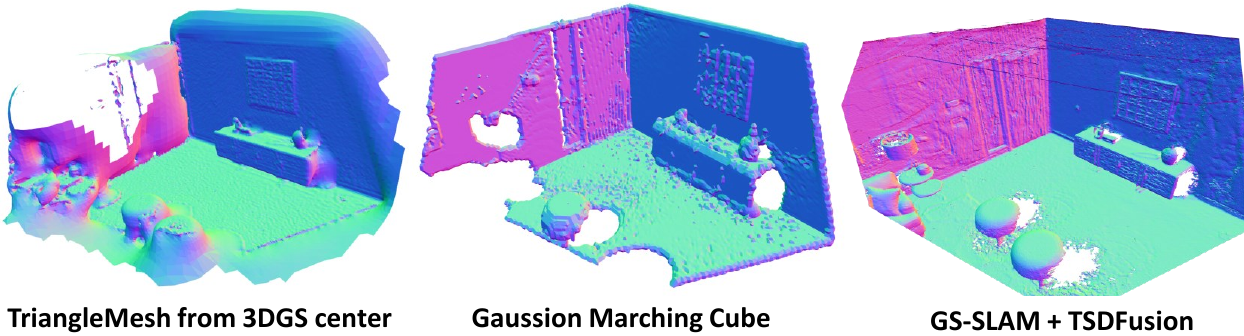}
    \end{center}
    \vspace{-3ex}
    \caption{Generate mesh from 3DGS with different methods.}
    \label{fig:gs_mcube}
\end{figure}

\boldparagraph{Mapping mesh comparison method.}
We follow Point-SLAM and use TSDF-Fusion~\cite{curless1996volumetric} to generate mesh from predicted pose and depth. We also evaluate map rendering quality in~\cref{sec:render}. That's because there is no direct way to get surface or mesh in 3DGS-based SLAM, as they do not represent scenes with density fields and can not directly generate mesh via marching cube. \ours achieves comparable map reconstruction results, better tracking accuracy, and higher FPS than Point-SLAM. Despite this, we explored generating mesh from 3DGS centers and Gaussian marching cube~\cite{Tang2023DreamGaussianGG} in~\cref{fig:gs_mcube}, but the results are not satisfactory. 

\boldparagraph{Tracking hyper-parameters.}
The pose is trained using FusedAdam optimizer with learning rate $lr_{\mathbf{t}}=2e^{-4}$, $lr_{\mathbf{q}}=5e^{-4}$, and photometric loss weighting 0.8, in the first five iterations we do the coarse pose estimation, while in the later iterations use the reliable 3D Gaussians to do the fine pose estimation. In addition, to exclude the pixel without proper color optimization. If the loss on the pixel is more than ten times the median loss, the pixel will be ignored.


\begin{figure*}[htbp]
    \vspace{-0ex}
    \centering
    {\footnotesize
        \setlength{\tabcolsep}{1pt}
        \renewcommand{\arraystretch}{0}
        \newcommand{\sz}{0.12}
        \begin{tabular}{cccc}
                                                                     & \texttt{\#fr1\_desk}                                                                                                          & \texttt{\#fr2\_xyz}                                                                                                          & \texttt{\#fr3\_office}                                                                                                          \\


            \rotatebox[origin=c]{90}{CoSLAM~\cite{Wang2023CoSLAMJC}} & \includegraphics[valign=c,height=\sz\linewidth]{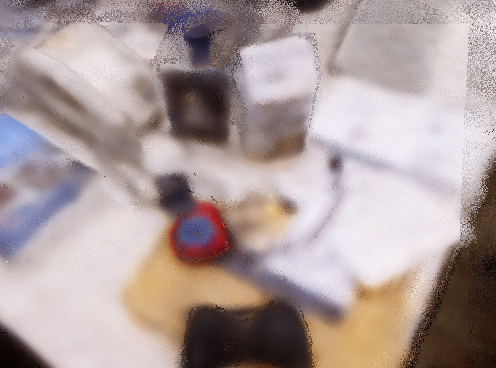}       & \includegraphics[valign=c,height=\sz\linewidth]{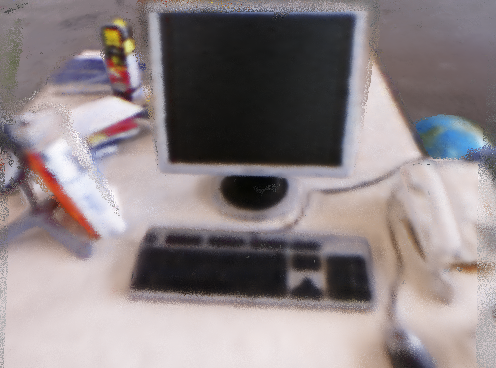}       & \includegraphics[valign=c,height=\sz\linewidth]{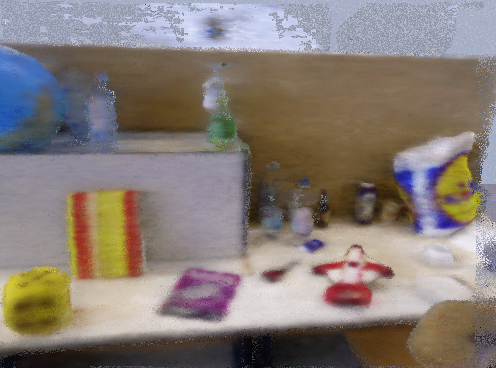}       \\

            \rotatebox[origin=c]{90}{ESLAM~\cite{Johari2022ESLAMED}} & \includegraphics[valign=c,height=\sz\linewidth]{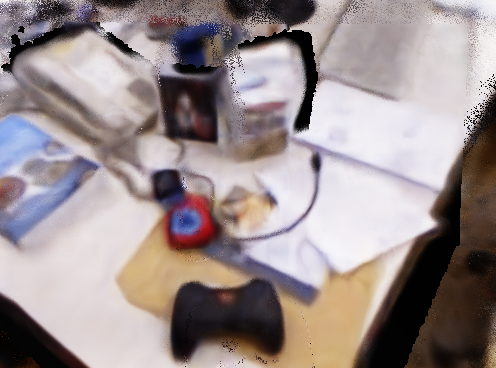}         & \includegraphics[valign=c,height=\sz\linewidth]{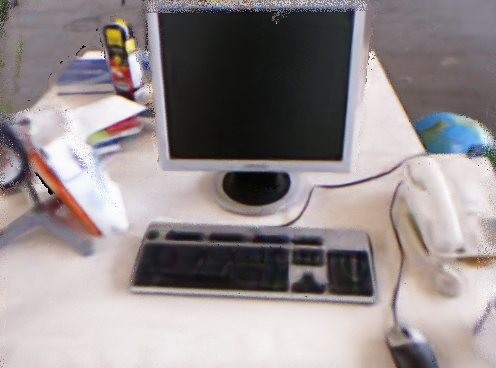}         & \includegraphics[valign=c,height=\sz\linewidth]{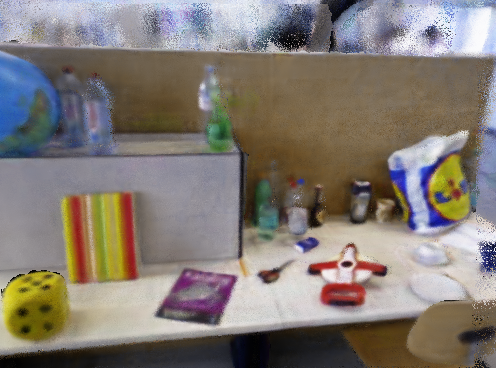}         \\

            \rotatebox[origin=c]{90}{Point-SLAM~\cite{pointslam}}    & \includegraphics[valign=c,height=\sz\linewidth]{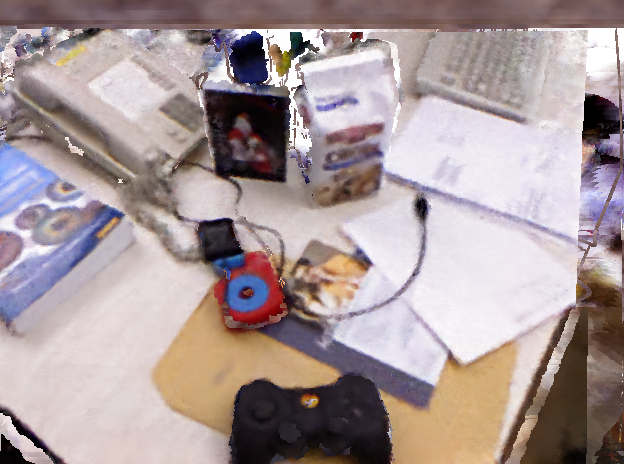} & \includegraphics[valign=c,height=\sz\linewidth]{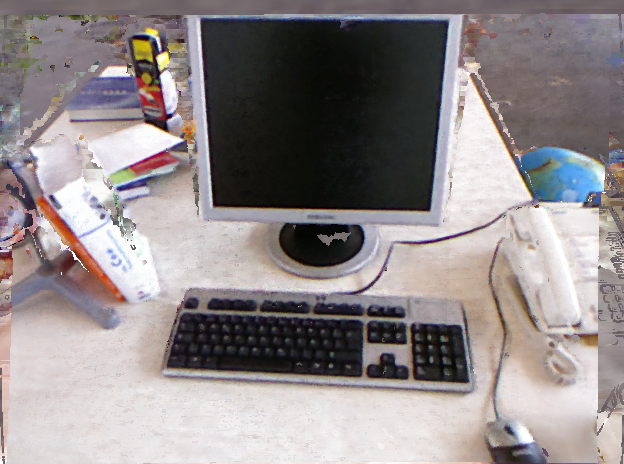} & \includegraphics[valign=c,height=\sz\linewidth]{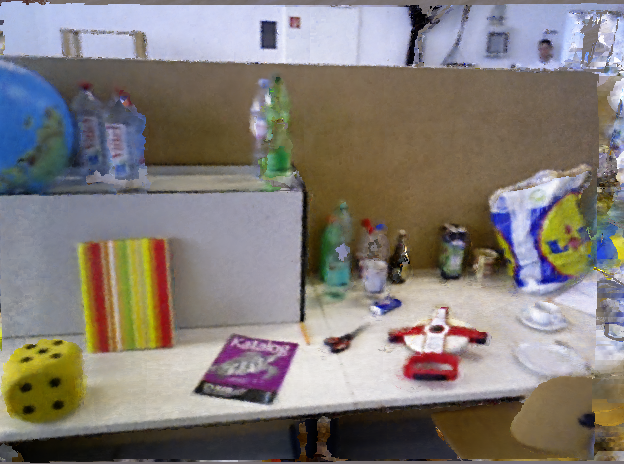} \\

            \rotatebox[origin=c]{90}{GS-SLAM (Ours)}                 & \includegraphics[valign=c,height=\sz\linewidth]{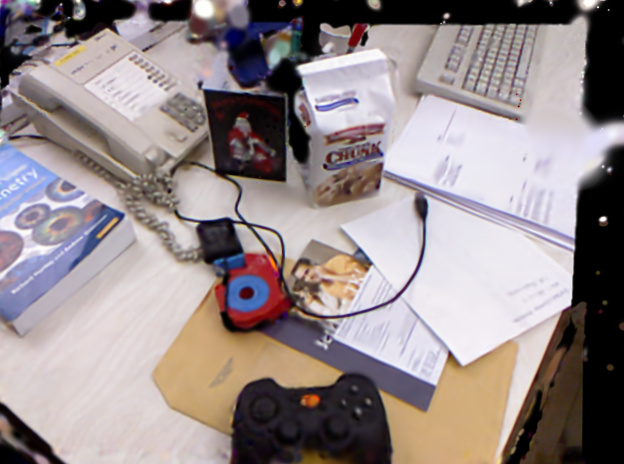}       & \includegraphics[valign=c,height=\sz\linewidth]{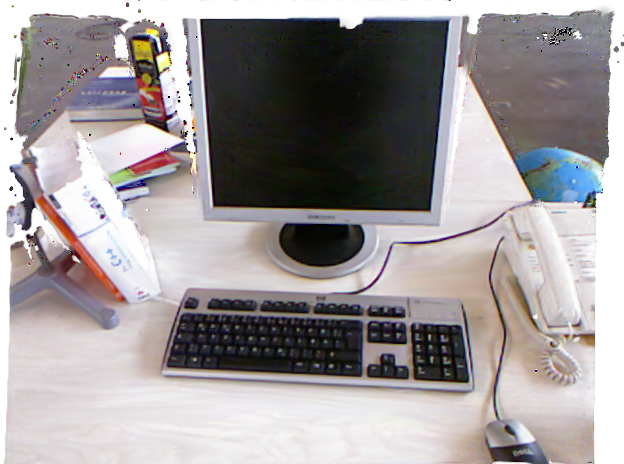}       & \includegraphics[valign=c,height=\sz\linewidth]{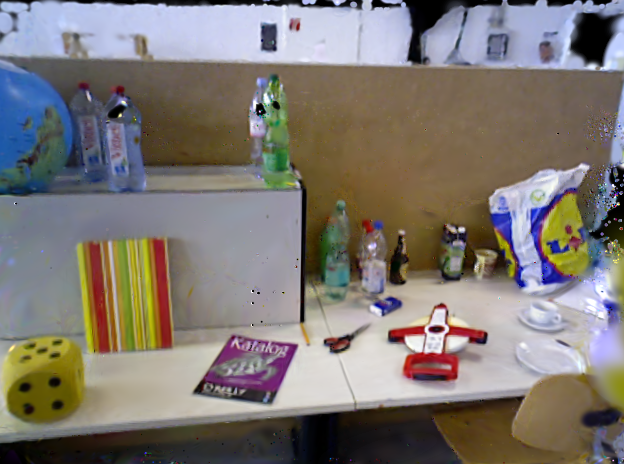}       \\

            \rotatebox[origin=c]{90}{Ground Truth}                   & \includegraphics[valign=c,height=\sz\linewidth]{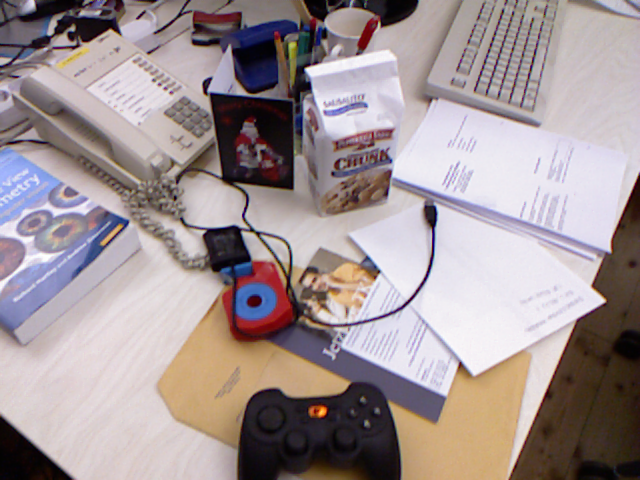}                 & \includegraphics[valign=c,height=\sz\linewidth]{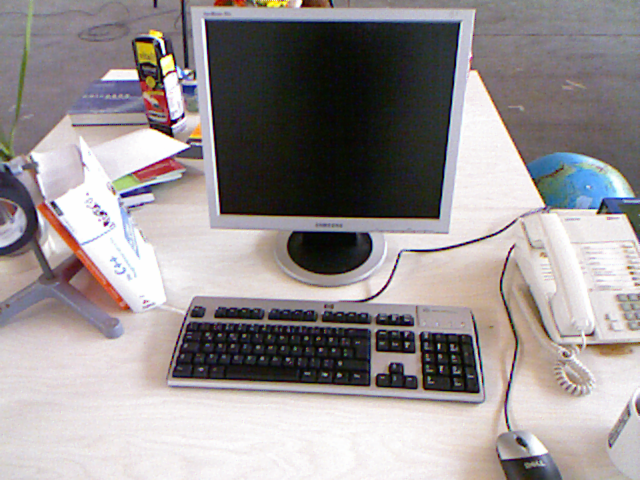}                 & \includegraphics[valign=c,height=\sz\linewidth]{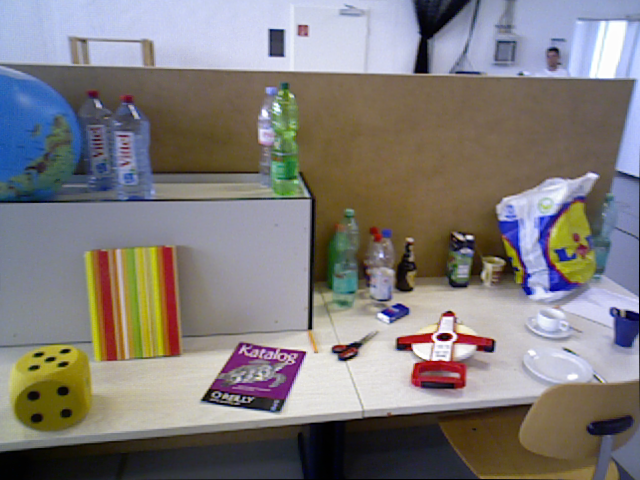}                 \\

        \end{tabular}
    }
    \caption{Additional render result comparison on TUM-RGBD~\cite{tum-rgbd_scribble_dataset}. Thanks to fast back-propagation of splatting in optimized 3D Gaussians, our GS-SLAM can reconstruct dense environment maps with richness and intricate details.}
    \label{fig:add_viz_tum}
\end{figure*}

\clearpage
\clearpage

\clearpage \clearpage
{
  \small
  \bibliographystyle{ieeenat_fullname}
  \bibliography{main}
}

\end{document}